\documentclass[10pt,twocolumn,letterpaper]{article}

\usepackage{iccv}
\usepackage{times}
\usepackage{epsfig}
\usepackage{graphicx}
\usepackage{amsmath}
\usepackage{amssymb}

\usepackage{color}
\usepackage{subfigure}
\usepackage{algorithm, algorithmic}
\usepackage{cite}
\usepackage{multirow}

\usepackage[utf8]{inputenc}
\usepackage{textcomp}
\usepackage{booktabs}  % for prettier tables
\usepackage{array} % for raggedleft tabular parapgrahs
\usepackage{lipsum} % HC for table alignment  
\usepackage{enumitem} % for no left margin in itemize
\usepackage{placeins}
\usepackage{diagbox}
\usepackage{makecell}

\usepackage[pagebackref=true,breaklinks=true,letterpaper=true,colorlinks,bookmarks=false]{hyperref}

\iccvfinalcopy % *** Uncomment this line for the final submission

\def\iccvPaperID{9197} % *** Enter the ICCV Paper ID here

\ificcvfinal\fi

\usepackage{chapterbib}

\begin{document}
    % \documentclass[10pt,twocolumn,letterpaper]{article}

% \usepackage{iccv}
% \usepackage{times}
% \usepackage{epsfig}
% \usepackage{graphicx}
% \usepackage{amsmath}
% \usepackage{amssymb}

% \usepackage{color}
% \usepackage{subfigure}
% \usepackage{algorithm, algorithmic}
% \usepackage{cite}
% \usepackage{multirow}

% \usepackage[utf8]{inputenc}
% \usepackage{textcomp}
% \usepackage{booktabs}  % for prettier tables
% \usepackage{array} % for raggedleft tabular parapgrahs
% \usepackage{lipsum} % HC for table alignment  
% \usepackage{enumitem} % for no left margin in itemize
% \usepackage{placeins}
% \usepackage{diagbox}

% % Include other packages here, before hyperref.

% % If you comment hyperref and then uncomment it, you should delete
% % egpaper.aux before re-running latex.  (Or just hit 'q' on the first latex
% % run, let it finish, and you should be clear).
% \usepackage[pagebackref=true,breaklinks=true,letterpaper=true,colorlinks,bookmarks=false]{hyperref}

% \iccvfinalcopy % *** Uncomment this line for the final submission

% \def\iccvPaperID{9197} % *** Enter the ICCV Paper ID here
% \def\httilde{\mbox{\tt\raisebox{-.5ex}{\symbol{126}}}}

% % Pages are numbered in submission mode, and unnumbered in camera-ready
% \ificcvfinal\pagestyle{empty}\fi

% \begin{document}

%%%%%%%%% TITLE
\title{Channel-Temporal Attention for First-Person Video Domain Adaptation}
% \title{Multi-Adversarial Channel-Temporal Attention for First-Person Video Domain Adaptation}
\author{Xianyuan Liu\textsuperscript{\rm 1,2,3}, 
Shuo Zhou\textsuperscript{\rm 3},
Tao Lei\textsuperscript{\rm 1,2},
Haiping Lu\textsuperscript{\rm 3}\\
\textsuperscript{\rm 1}Institute of Optical and Electronics, Chinese Academy of Sciences, China\\
\textsuperscript{\rm 2}University of Chinese Academy of Sciences, China\\
\textsuperscript{\rm 3}University of Sheffield, United Kingdom\\
\{xianyuan.liu, szhou20, h.lu\}@sheffield.ac.uk, taoleiyan@ioe.ac.cn
}

% \and

\maketitle
\ificcvfinal\thispagestyle{empty}\fi

%%%%%%%%% ABSTRACT
\begin{abstract}
Unsupervised Domain Adaptation (UDA) can transfer knowledge from labeled source data to unlabeled target data of the same categories. However, UDA for first-person action recognition is an under-explored problem, with lack of datasets and limited consideration of first-person video characteristics. This paper focuses on addressing this problem. Firstly, we propose two small-scale first-person video domain adaptation datasets: ADL$_{small}$ and GTEA-KITCHEN. Secondly, we introduce channel-temporal attention blocks to capture the channel-wise and temporal-wise relationships and model their inter-dependencies important to first-person vision. Finally, we propose a 
Channel-Temporal Attention Network (CTAN) to integrate these blocks into existing architectures. CTAN outperforms baselines on the two proposed datasets and one existing dataset EPIC$_{cvpr20}$.
% Unsupervised Domain Adaptation (UDA) can transfer knowledge from labeled source data to unlabeled target data where annotation is not available and impractical. % (to (reduce the need of annotation.)).
% However, few attempts have been devoted to UDA for first-person action recognition due to the lack of datasets,
% % and the special characteristics.
% resulting in few exploration in the special characteristics of first-person video for UDA. This paper focuses on addressing this problem. First, we propose two small-scale video domain adaptation datasets: ADL$_{small}$ and GTEA-KITCHEN. Second, we introduce channel-temporal attention blocks to capture the channel-wise and temporal-wise relationships and model their inter-dependencies. Finally, we propose a 
% % Multi-Adversarial 
% Channel-Temporal Attention Network (CTAN) to integrate these blocks into the existing architecture. Our network outperforms the baselines on our proposed datasets and EPIC$_{cvpr20}$.
\end{abstract}

%%%%%%%%% BODY TEXT
\section{Introduction}
% \input{sections/introduction}
% Action recognition is a fundamental and core problem in computer vision and has been widely studied. Motivated by the success of deep learning techniques like Convolutional Neural Networks (CNNs) on image classification, recent studies \cite{Feichtenhofer_2016_CVPR,Carreira_2017_CVPR,tran2018closer,8454294} have focused on CNN-based architectures to recognize the action and achieved state-of-the-art performance on several benchmarks \cite{Kuehne11,soomro2012ucf101,kay2017kinetics,Damen2018EPICKITCHENS}. However, in many practical applications, the labeled dataset (source domain) and unlabeled novel dataset (target domain) exists mismatch, result in the distribution mismatch of the training set and test set, which is known as \textit{domain shift}. These two domains are related but not in the same distribution, and mismatch tend to result in a poor performance on action performance. As shown in Table \ref{tab:ADL_res}, domain discrepancy not only exists between different datasets but also between different videos in the same dataset. 
% One solution is fine-tuning the existing network on the novel dataset to fit the new domain. However, it requires annotating all data for the novel dataset, which may be impractical because it is expensive and time-consuming to collect enough labeled data to properly train a network with such a considerable number of parameters. 
Video-based action recognition is a challenging computer vision problem. Great progress has been made on several benchmark datasets \cite{Kuehne11,soomro2012ucf101,kay2017kinetics,Damen2018EPICKITCHENS} using architectures based on Convolutional Neural Networks (CNNs) \cite{Feichtenhofer_2016_CVPR,Carreira_2017_CVPR,tran2018closer,8454294}. However, it remains a major obstacle to generalize models learned on one domain to another domain due to the distribution mismatch between them, i.e., the \textit{domain shift}. Moreover, domain discrepancy not only exists between different datasets but also between different videos in the same dataset, e.g. as shown in Table \ref{tab:ADL_res}.

\begin{figure}[ht]
 \centering
\includegraphics[width=6cm]{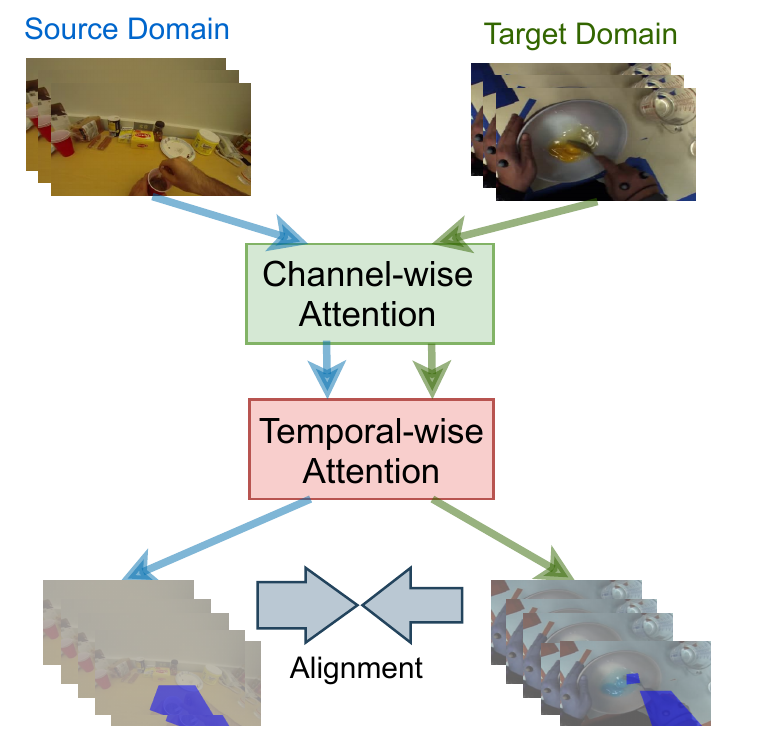}
\caption{First, we learn common channel-wise and temporal-wise attention for both source and target videos to focus on interactions important for actions in first-person videos (e.g. by the hands). Then we adapt the network to align source and target distributions.}
% \caption{The Overview of proposed Channel-Temporal Attention Network (CTAN) for first-person video UDA. Actions always occur around hands in first-person videos (purple areas). Therefore, we propose to make the network focus on these areas while domain adaptation. Here we use the action \emph{pour} as the example.}
\label{fig:overall_figure}
\end{figure}

Unsupervised Domain Adaptation (UDA) can address the domain shift problem by designing networks that take domain discrepancy into account to learn common features. Such networks can explore a shared distribution for the two domains without labels from the target domain. Recently, UDA has shown remarkable progress on still images such as object detection \cite{Kim_2019_ICCV}, person re-identification \cite{Fu_2019_ICCV}, and semantic segmentation \cite{Jaritz_2020_CVPR}. While UDA methods for still images focus on minimizing the distribution distance between domains in the spatial feature space \cite{Yan_2017_CVPR,sun2016return,ganin2016domain,long2018conditional}, UDA methods for videos aim to minimize the distribution distance between domains in both spatial and temporal feature spaces.

% the network can explore a shared distribution for the two domains without labels from target domain data. Recently, UDA has shown remarkable progress on still images such as object detection \cite{Kim_2019_ICCV}, person re-identification \cite{Fu_2019_ICCV} and semantic segmentation \cite{Jaritz_2020_CVPR}. Compared with still image, which only need to focus on minimizing the distribution distance between different domains in the spatial feature space \cite{Yan_2017_CVPR,sun2016return,ganin2016domain,long2018conditional}, videos need to take both spacial and temporal feature into consideration. Therefore, UDA on action recognition requires matching the distributions between domains in both spatial and temporal feature spaces.

% \begin{figure}[t]
%  \centering
% \includegraphics[width=5cm]{figure1.png}
% \caption{The recognition accuracy across videos in ADL datasets. The performance changes sharply when testing on the same dataset but training on different dataset. It shows different videos in the same dataset can have domain discrepancy.}
% \label{fig:age_mmse}
% \end{figure}

% \multicolumn{2}{c}{Accuracy} & \multicolumn{3}{c}{\textbf{Train}} \\
\begin{table}[t]
% \vspace{-2mm}
\centering{
%\begin{small}
  \begin{tabular}{c  c c c c}
\toprule
Test  \textbackslash Train  & Video 1 & Video 2 & Video 3\\
\midrule
Video 1 & 95.8 & 25.0 & 27.4 \\
%\midrule
Video 2 & 41.1 & 93.5 & 37.5 \\
% \midrule
Video 3 & 28.6 & 24.8 & 95.1 \\
\bottomrule
\end{tabular}
%\end{small}
 \vspace{1mm}
    \caption{Action recognition accuracy (\%) across three videos from the same ADL dataset \cite{pirsiavash2012detecting}. }
    }
    \label{tab:ADL_res}
\end{table}
%Even if training on the same video, the network performs sharply change when testing on different video.

We can categorize existing video datasets for action recognition into third-person vision and first-person vision datasets \cite{ryoo2013first}. This paper focuses on UDA for first-person action recognition, where videos are collected by a wearable camera so the availability is limited compared to the abundant third-person datasets. 
To the best of our knowledge, UDA for first-person action recognition has only been studied on subsets of the EPIC-KITCHEN (EPIC) dataset \cite{Damen2020RESCALING}, where the first version in \cite{munro20multi} (EPIC$_{cvpr20}$) took 9 hours to train on 8 V100 GPUs with a batch size of 128. For researchers with limited computing resource, it is difficult/infeasible to develop and study UDA models on such a large-scale dataset. This motivates us to build datasets that are of smaller scale yet still meaningful for UDA to lower the bars and accelerate the development in this area. Therefore, we first propose two small-scale, first-person action recognition datasets for UDA: 
1) ADL$_{small}$: We collect three long-duration videos from the ADL dataset \cite{pirsiavash2012detecting} and restructure them for UDA. 2) GTEA-KITCHEN: We combine two first-person video datasets, GTEA \cite{fathi2011learning} and KITCHEN \cite{de2009guide}, for UDA by restructuring the GTEA dataset labels and re-label the KITCHEN dataset manually. These two datasets are small-scale but provide large domain shift for UDA research. See more details in Section \ref{sec:uda_dataset} and Table \ref{tab:gtacc} and Table \ref{tab:adlacc}. 

There are two categories of UDA networks for action recognition, with different temporal feature extraction mechanisms. The first category \cite{munro20multi} takes a two-stream approach to 1) extract spatial information and temporal information separately from video data, 2) transfer spatial and temporal knowledge separately between datasets, and 3) fuse both in the end. For example, the MM-SADA method \cite{munro20multi} utilizes RGB frame and optical flow as the two streams to boost UDA performance on an EPIC subset (EPIC$_{cvpr20}$). However, it has a high space and time complexity. Moreover, although MM-SADA includes a self-supervised module to capture the relationship between RGB frames and optical flow, the extraction processes for spatial and temporal information are separated. 

The second category extracts spatio-temporal information from videos directly and transfers spatio-temporal knowledge between domains \cite{Chen_2019_ICCV,jamal2018deep}. $\rm TA^3N$ \cite{Chen_2019_ICCV} extends image-based UDA to video by adding an attentive temporal alignment to increase UDA performance on third-person videos. However, $\rm TA^3N$ needs features extracted by 2D CNN as the input and does not extract spatio-temporal features from videos directly, which could not learn task-specific features in an end-to-end fashion. Moreover, the attention mechanism in $\rm TA^3N$ is for the extraction and adaptation of temporal domain features only, rather than spatio-temporal information.

Compared with third-person videos, first-person videos have some unique characteristics. For example, actions tend to occur in some local areas, and particularly where the hands and objects interact. Thus, we hypothesize that networks paying more attention to such areas can leverage such characteristics to benefit common feature learning in UDA, as shown in Figure \ref{fig:overall_figure}. In addition, different channels in different layers of CNNs capture different characteristics. The Squeeze-Excitation (SE) block in \cite{hu2018squeeze} can generate attention scores to excite important channels. This inspires us to design excitation attention approaches that can weigh the channel-wise and temporal-wise features in the CNN layers to reveal the channel-temporal relationships for first-person videos. To this end, we propose a Channel-Temporal Attention block (CTA) that can make the network to pay more attention to action-related features in first-person videos. Moreover, for UDA, the network should not only focus on these important features but also focus on the common features across domains. Therefore, we utilize an adversarial approach at the video level for alignment to minimize the discrepancy between important channels in the source and target domains.

In summary, our contributions are three-fold:
\begin{itemize}
    \item [1)]
    \textit{First-person Video Domain Adaptation Dataset Collection}: We collect two small-scale first-person datasets for UDA, ADL$_{small}$ and GTEA-KITCHEN. They both have sufficient domain shift and can lower the bars for researchers to enter this field, stimulate more studies, and accelerate research in this area. To our knowledge, these two datasets are the only datasets besides the EPIC dataset \cite{munro20multi, Damen2020RESCALING} for studying the first-person video UDA problems.
\end{itemize}
\begin{itemize}
    \item [2)]
    \textit{Channel-Temporal Excitation Attention for First-person Video UDA}: We explore different excitation attention approaches for UDA by modeling channel-wise and temporal-wise inter-dependencies. The results show that these approaches can make the network better focus on the key features from first-person videos and benefit the video UDA.
\end{itemize}
\begin{itemize}
    \item [3)]
    \textit{Channel-Temporal Attention Network}: We propose a new adversarial Channel-Temporal Attention Network (CTAN) with the channel-temporal excitation attention above. We train our network end-to-end based on I3D \cite{Carreira_2017_CVPR} and test on our proposed small-scale datasets and the large-scale EPIC$_{cvpr20}$ dataset. Our network outperforms all the baselines on the three datasets.
    % [[[Our network achieves state-of-the-art performance on both small-scale and large-scale first-person video DA datasets.]]]
\end{itemize}

\section{Related Works}
\begin{figure*}[ht]
 \centering
\includegraphics[width=15cm]{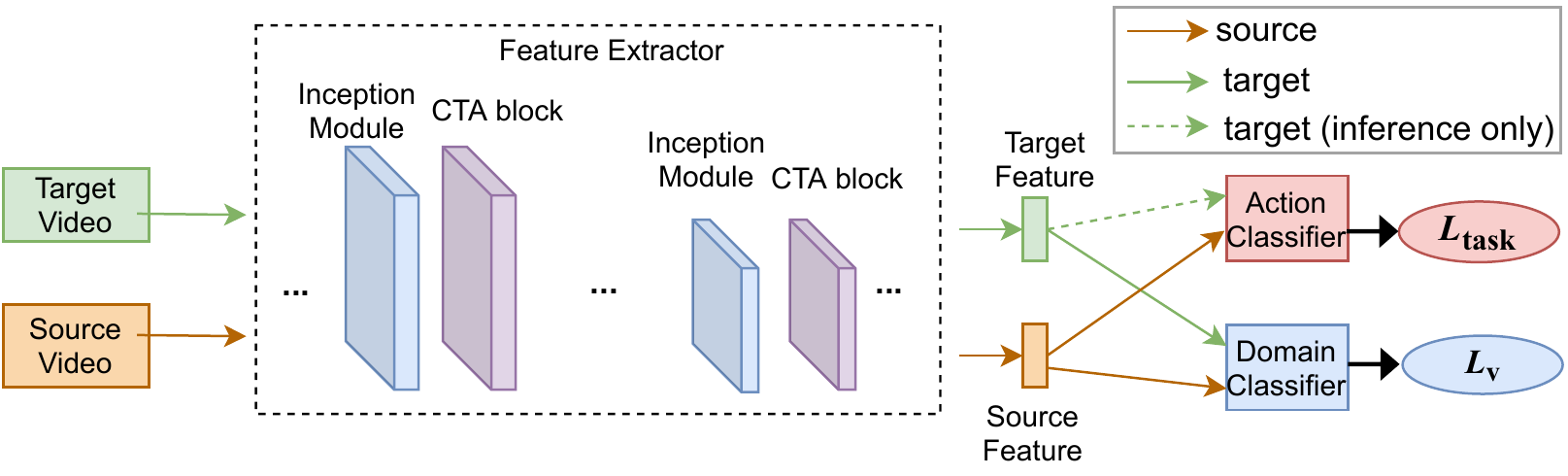}
\caption{The proposed Channel-Temporal Attention Network (CTAN) model for first-person action recognition. Source and target domains share the same feature extractor, which is composed of Inception modules \cite{Carreira_2017_CVPR} and proposed Channel-Temporal Attention (CTA) blocks. In training, the feature extractor takes labeled source videos and unlabeled target videos as the input and generates source and target features as the output. Source features are fed into both action and domain classifiers, while target features are only fed to domain classifier. In test, only target videos are the input to the feature extractor and then the action classifier.}
% \caption{The framework of the proposed network for UDA in first-person action recognition. The feature extractor, shared for both source and target domains, is composed of Inception modules \cite{Carreira_2017_CVPR} is shared for both source and target domains and Channel-Temporal Attention CTA blocks. In the training, the inputs of the feature extractor are videos from labeled source and unlabeled target domains and the outputs are source and target feature. Source feature is fed into both task and domain classifiers, while target feature is only applied to domain classifier. During test, only target videos are the inputs and tested on feature extractor and action classifier.}
\label{fig:main_structure}
\end{figure*}

\subsection{Action Recognition} 
There are three categories of action recognition networks according to different ways to extract temporal features. The first category takes a two-stream approach to extract temporal features, typically optical flow, directly and then combine them with spatial feature, via late fusion \cite{simonyan2014two}, constant fusion \cite{Feichtenhofer_2016_CVPR}, and sparse sampling \cite{8454294}.

The second category takes spatial features as inputs to extract spatio-temporal features directly by the 3D CNNs. C3D \cite{tran2015learning} constructs 3D kernels to extract short-term information from the RGB frame input. R(2+1)D \cite{tran2017convnet} applies skip connection to C3D and explore different 3D and 2D convolution combinations. I3D \cite{Carreira_2017_CVPR} inflates 2D convolutional and pooling kernels of 2D CNN trained on image datasets into 3D to use well-trained 2D CNN parameters. In general, I3D is considered to be superior to C3D and R(2+1)D.

The third category utilizes temporal modeling to extract spatio-temporal feature from spatial inputs, via recurrent neural networks \cite{7558228}, multi-scale temporal relation pooling \cite{zhou2017temporalrelation}, or spatial feature channel shifting \cite{lin2019tsm}.

We choose I3D as our backbone considering the trade-off between performance and model size. However, our proposed method is applicable to the two-stream and temporal modeling networks as well, with potentially better performance but higher computational cost.%, while C3D lacks performance and R(2+1)D have much more parameters than I3D.

\subsection{Unsupervised Domain Adaptation} 

Unsupervised domain adaptation aims to find a common feature space between the labeled source data and unlabeled target data, with three approaches. The first approach is discrepancy-based. It aligns source-target distributions by minimizing a divergence that measures the distance between them, e.g. via weighted Maximum Mean Discrepancy (MMD) \cite{Yan_2017_CVPR} or Correlation Alignment (CORAL) \cite{sun2016return}.% are basic metric for the distance. 

The second approach is adversarial-based. It utilizes domain discriminators and conducts adversarial training to reduce the discrepancy. DANN \cite{ganin2016domain} utilizes discriminators and gradient reversal layer (GRL) to accomplish alignment through standard back-propagation training. CDAN \cite{long2018conditional} leverages multilinear and entropy conditioning on discriminative information to enable alignment of multi-modal distributions.

The third approach is reconstruction-based. DRCN \cite{ghifary2016deep} uses a pair-wise squared reconstruction loss to reconstruct the target data, while DSN \cite{bousmalis2016domain} uses scale-invariant mean squared error reconstruction loss to reconstruct both the source and target data. 

%Considering the huge difficulty of applying reconstruction to video, 
We consider the first two approaches, specifically, their extensions from image UDA to video UDA.

\subsection{Domain Adaptation for Action Recognition} 

Most domain adaptation models for action recognition consider third-person videos. DAAA \cite{jamal2018deep} utilizes 3D CNN to extract spatio-temporal feature, projects them to a latent subspace, and then uses discriminators to reduce the discrepancy in the subspace. TA$^3$N \cite{Chen_2019_ICCV} utilizes temporal relation module from \cite{zhou2017temporalrelation} to extract spatio-temporal features and extends image-based domain adaptation to videos by adding temporal attentive alignment. TCoN \cite{pan2020adversarial} applies a co-attention mechanism to CNN features to guide the network to focus on common keyframes across domains on RGB and optical flow. %However, they all focus on third-person videos rather than first-person videos. 

For domain adaptation on first-person videos, MM-SADA \cite{munro20multi} adopts two-stream networks and utilizes self-supervised multi-modal UDA to learn the relationship between RGB and optical flow. It has high computational cost due to the usage of optical flow for creating multi-modal inputs. There is no consideration of unique characteristics of first-person videos either

In contrast to the above approaches, this paper will utilize the excitation attention approach to take the characteristics of first-person videos into account in a channel-wise and temporal-wise manner.

% \newpage
\section{Proposed Method}
Figure \ref{fig:main_structure} shows the proposed unsupervised domain adaptation model for first-person action recognition, named as Channel-Temporal Attention Network (CTAN). In training, source and target videos are fed into a feature extractor that modifies the I3D \cite{Carreira_2017_CVPR} pretrained on Kinetics by adding multiple channel and temporal attention (CTA) blocks. Each proposed CTA block contains a channel attention module and a temporal attention module and it is inserted into I3D to perform re-calibration of both channel-wise and temporal-wise features. After feature extraction, source features are fed into an action  classifier and also, both source and target features are fed into a  discriminator for adversarial domain discrimination. In test, only target videos are taken as the input to the feature extractor to extract features for the action classifier to predict the action class.

\subsection{Channel-Temporal Attention Module} 
\begin{figure}[t]
\subfigure[Channel-wise attention module]{
    \begin{minipage}[t]{0.5\linewidth}
        \centering
        \includegraphics[width=1.6in]{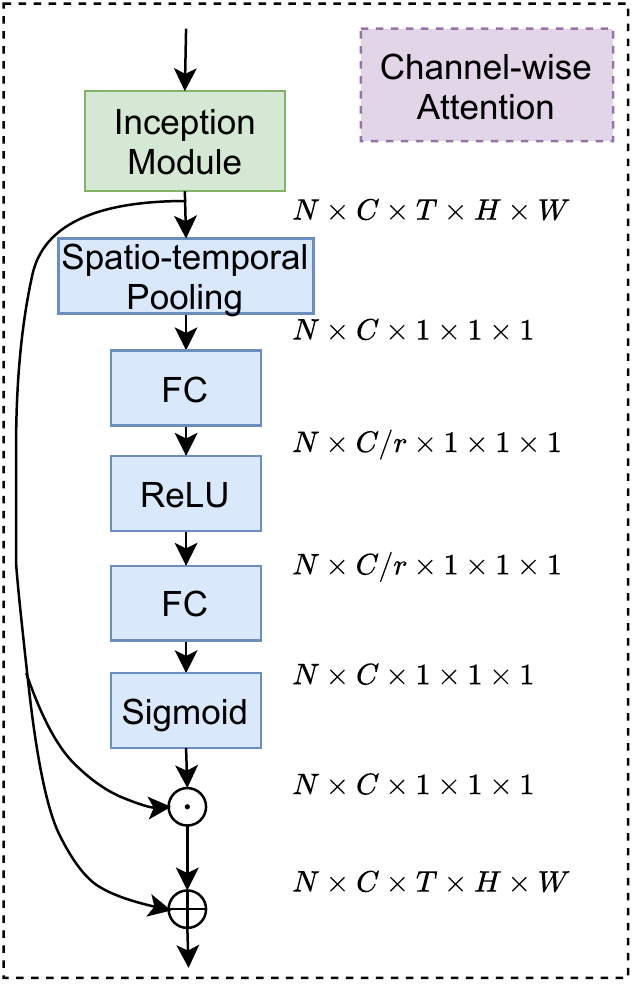}
        \label{fig:a}
    \end{minipage}%
}%
\subfigure[temporal-wise attention module]{
    \begin{minipage}[t]{0.5\linewidth}
        \centering
        \includegraphics[width=1.6in]{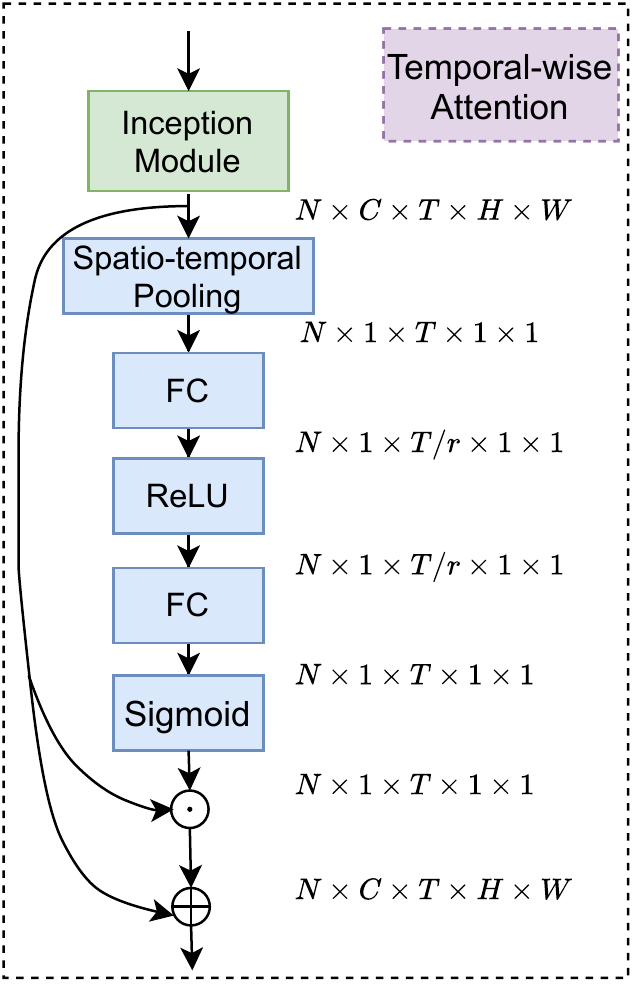}
        \label{fig:b}
    \end{minipage}%
}%
\centering
  \setlength{\abovecaptionskip}{-0.2cm}
  \setlength{\belowcaptionskip}{-0.8cm} 
  \vspace{-0.2cm}
% \caption{The implementations of the proposed chan.}
\caption{Architecture of the channel-temporal attention module.}
\label{fig:ctblock}
\end{figure}

In first-person video action recognition, different channels of CNN layers capture different spatio-temporal information related to the action. Such spatio-temporal information can benefit domain adaptation for action recognition.
% Taking the use of the critical information while feature learning and DA between the two domains can benefit the performance.
Firstly, inspired by the SE block \cite{hu2018squeeze} that excites informative features in input image channels, we extend the SE block to channel-wise attention (CA) module for video input, as shown in Figure \ref{fig:ctblock}. Secondly, people can usually recognize an action at a glance as long as they see a small but informative part of this action. This phenomenon inspires us to extend the previous CA module to the temporal-wise attention (TA) module so that our network can focus on informative features in temporal dimensions. Finally, we integrate the CA and TA modules into the channel-temporal attention (CTA) module described in detail below.

Given a 5D video feature $ \mathbf{X} \in \mathbb{R}^{N \times T \times C \times H \times W}$. $N$, $T$ and $C$ denote batch size, temporal dimension and feature channel size. $H$ and $W$ correspond to height and width. First, we utilize 3D average pooling to extract channel-wise information among dimensions $T$, $H$ and $W$. Then, we capture the channel-reduced feature for efficiency by a linear layer with parameters $\mathbf{W}_{c1}$ and a reduction ratio $r$.
% \begin{equation}
%     \textbf{X}_c = \frac{\mathrm{1}}{T\times H\times W}\sum_{t=1}^T\sum_{h=1}^H\sum_{w=1}^W \textbf{X}_{t,h,w} \label{eq:poolc}
% \end{equation}
% \begin{equation}
%     X_c = ReLU({W}_{c1}(\textbf{X}_c)) \label{eq:fc}
% \end{equation}
\begin{equation}
    \mathbf{X}_c = \frac{\mathrm{1}}{T\times H\times W}\sum_{t=1}^T\sum_{h=1}^H\sum_{w=1}^W \mathbf{X}_{t,h,w} \label{eq:poolc}
\end{equation}
\begin{equation}
    \mathbf{X}_{cr} = \mathrm{ReLU}(\mathbf{W}_{c1}\mathbf{X}_c) \label{eq:xcr}
\end{equation}
where $\mathbf{X}_c \in \mathbb{R}^{N \times 1 \times C \times 1 \times 1}$ denotes the output of pooling and $\mathbf{X}_{cr} \in \mathbb{R}^{N \times 1 \times C/r \times 1 \times 1}$ denotes the squeezed channel-reduced feature.

Another linear layer is used with parameters $\mathbf{W}_{c2}$ to restore the channel dimension of the feature and a sigmoid function $\sigma$ is used to capture channel-attentive weights $\mathbf{A}_c \in \mathbb{R}^{N \times 1 \times C \times 1 \times 1}$. In order to excite the informative channels, we compute a Hadamard product $\odot$ between these weights $\mathbf{A}_c$ and the video feature $\mathbf{X}$ as
\begin{equation}
    \mathbf{X}_c^o = \mathbf{X}+\mathbf{A}_c\odot \mathbf{X} = \mathbf{X}+\sigma(\mathbf{W}_{c2}\mathbf{X}_{cr})\odot \mathbf{X}
\end{equation}
where $\mathbf{X}_c^o \in \mathbb{R}^{N \times T \times C \times H \times W}$ denotes the output of the channel attention module with the excited and enhanced channel-wise informative features. Because wrong channel attention may hurt the performance to some degree and some channel attention may suppress other information, we add a residual connection to mitigate these negative effects.

We then feed the output $\mathbf{X}_c^o$ into the temporal module and conduct a 3D average pooling on the video feature $\mathbf{X}_c^o$ among $C$, $H$ and $W$ to extract the temporal-wise information. We adopt two linear layers to model the temporal feature. 
% Note that we do not use reduction here because the temporal dimension is usually small in the CNNs.
The sigmoid function $\sigma$ is again adopted to get the temporal-attentive weights $\mathbf{A}_t\in \mathbb{R}^{N \times T \times 1 \times 1 \times 1}$. The Hadamard product $\odot$ is computed to excite temporal features on the input $\mathbf{X}_c^o$. A skip connection is also applied to prevent temporal attention from suppressing other information.
% \begin{equation}
%     X_t = \sigma W_{t1}(\frac{\mathrm{1}}{C\times H\times W}\sum_{c=1}^C\sum_{h=1}^H\sum_{w=1}^W {X_c^o}_{c,h,w} \label{eq:pool2})
% \end{equation}

\begin{equation}
    \mathbf{X}_t = \frac{\mathrm{1}}{C\times H\times W}\sum_{c=1}^C\sum_{h=1}^H\sum_{w=1}^W {\mathbf{X}_c^o}_{c,h,w} \label{eq:pool2}
\end{equation}
\begin{equation}
    \mathbf{X}_{tr} = \mathrm{ReLU}(\mathbf{W}_{t1}\mathbf{X}_t) \label{eq:xtr}
\end{equation}
\begin{equation}
    \mathbf{X}_{ct}^o = \mathbf{X}_c^o+\mathbf{A}_t\odot\mathbf{X}_c^o = \mathbf{X}_c^o+\sigma(\mathbf{W}_{t2}\mathbf{X}_{tr})\odot{\mathbf{X}_c^o}
\end{equation}
where $\mathbf{X}_t\in\mathbb{R}^{N \times T \times 1 \times 1 \times 1}$ and $\mathbf{X}_{tr}\in\mathbb{R}^{N \times T/r \times 1 \times 1 \times 1}$ denote the squeezed temporal feature and temporal-reduced feature. $\mathbf{W}_{t1}$ and $\mathbf{W}_{t2}$ are the parameters of the linear layer. $\mathbf{X}_{ct}^o \in \mathbb{R}^{N \times T \times C \times H \times W}$ is the excited output.

\subsection{Adversarial UDA}
\label{sec:adv_uda}
After introducing the CTA for important channel-temporal feature extraction from source domain, we need to consider how to design the network for learning the common features across domains. For UDA, the network needs to learn common features across domains while focusing on the important features for target domain classification. For convenience, discrepancy-based and adversarial-based approaches such as DAN \cite{long2015learning} and DANN \cite{ganin2016domain} are easy to be adapted to our task compared with reconstruction-based UDA.
% approaches, which need to reconstructing videos. 
In comparison, linear version DAN needs a large batch size to avoid negative MMD loss resulting in more computation resource requirement than DANN.
% Previous networks for UDA in first-person action recognition \cite{jamal2018deep,Chen_2019_ICCV,munro20multi} normally utilizes the domain discriminator to bridging the distribution discrepancy between source and target domain data. However, these networks focus on video-level feature rather than the more detail feature like channel-level feature. Therefore, we propose a multi-adversarial approach to align both video-level and channel-level features.

In this paper, we utilize the domain adversarial neural network (DANN) \cite{ganin2016domain}, in which a two-player mini-max game is constructed considering the limited computation resources. The main idea of DANN is to add one domain classifier $G_d$ to discriminate whether the data is from the source or target domain. The parameters of the domain classifier are trained by minimizing the discriminator loss $L_d$, while another feature extractor $G_f$ maximizes this discriminator loss to train the extractor parameters. The aim is to confuse the discriminator to guide the feature extractor to learn common features between the source and target domain. Here, we utilize a discriminator as in DANN to align features extracted by the feature extractor across domains. The domain loss $L_v$ is defined for each video input $x_i$ as:
\begin{equation}
    L_v=-\frac{1}{n}\sum_{x_i\in D_s \cup D_t }L_d(G_d(G_f(x_i)), d_i)
\end{equation}
where $D_s$ and $D_t$ are source and target domains respectively, $n$ is the number of samples from both domains, and $d_i$ is the domain label of $x_i$. If $x_i$ is from the source (target) domain, $d_i$ is set as 1 (0)
% , while $d_i$ is 0 if $x_i$ is from target domain.

% We then add discriminators into our CTA modules to align channel-level features and calculate the channel-wise domain loss $L_c$ as follow:
% \begin{equation}
    % L_c=-\frac{1}{Kn}\sum_{k=1}^{K}\sum_{x_i\in D_s \cup D_t}L_d(G_d^k(G_b^k(x_i)), d_i)
% \end{equation}
% where $G_b^k$ is the feature extractors for channels in different layers and $G_d^k$ refers to channel-level discriminators. $K$ is the number of $G_d^k$.

\subsection{Integration with I3D Network}
Finally, we integrate the proposed modules and adversarial UDA into I3D, as illustrated in Figure \ref{fig:main_structure}. The channel-temporal attention modules are integrated into I3D after Inception modules. Following the finding in \cite{hu2018squeeze} that lower layer features are typically more general, while higher layer features have greater specificity, we integrate our proposed modules into Inception module ``Mixed4b" to Inception module ``Mixed4f" in the I3D architecture instead of the too early and too late modules.
% and channel-level discriminators are inserted after the output of the CA module. 
The adversarial discriminator and a two-layer action classifier $G_y$ are integrated after the average pooling layers of I3D. Gradient reversal layer (GRL) is also used between the discriminator and feature extractor to invert the gradient. The overall loss $L$ can be expressed as follows:
\begin{equation}
    \begin{aligned}
    %  L=& L_{task}+\lambda_v L_v+\lambda_c L_c \\
    L=& L_{task}+\lambda_v L_v \\
    =&\frac{1}{n_s}\sum_{x_i\in D_s}L_y(G_y(G_f(x_i)), y_i)\\
    -&\frac{\lambda_v}{n}\sum_{x_i\in D_s \cup D_t }L_d(G_d(G_f(x_i)), d_i)\\
    % -&\frac{\lambda_c}{Kn}\sum_{k=1}^{K}\sum_{x_i\in D_s \cup D_t}L_d(G_d^k(G_b^k(x_i)), d_i)
    \end{aligned}
\end{equation}
% where $\lambda_v$ and $lambda_c$ are the hyper-parameter to trade-off domain adaptation with classification respectively. $y_i$ refers to the action labels of input $x_i$.
where $\lambda_v$ is a hyper-parameter to trade-off domain adaptation with classification respectively. $y_i$ refers to the action labels of input $x_i$. The whole network is trained by two cross entropy loss, $L_y$ and $L_d$. 

% \newpage
% test
% \newpage
\section{Proposed First-Person UDA Datasets}
\label{sec:uda_dataset}

\begin{figure}[t]
\subfigure[ADL$_{small}$]{
    \begin{minipage}[t]{0.5\linewidth}
        \centering
        \includegraphics[width=1.75in]{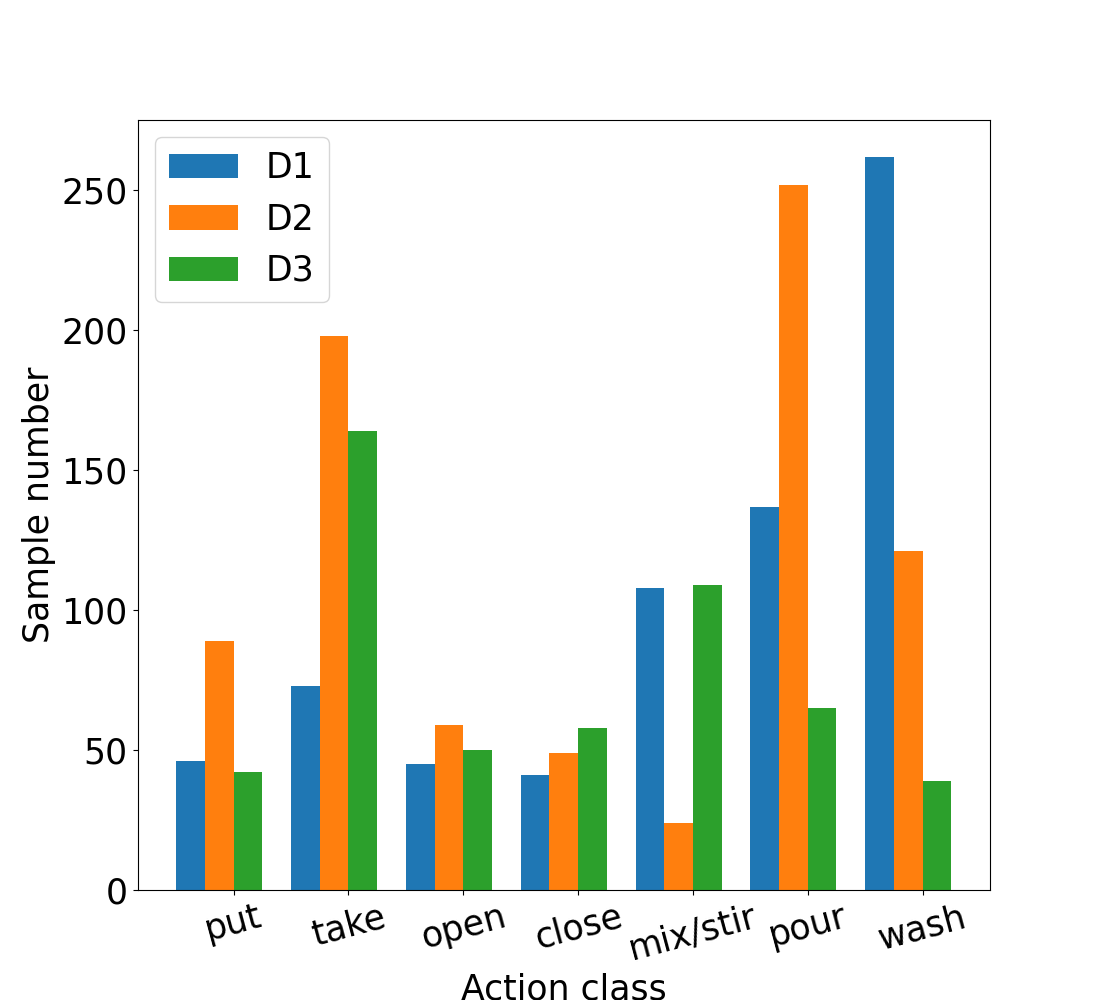}
        \label{fig:adl_class}
    \end{minipage}%
}%
\subfigure[GTEA-KITCHEN]{
    \begin{minipage}[t]{0.5\linewidth}
        \centering
        \includegraphics[width=1.75in]{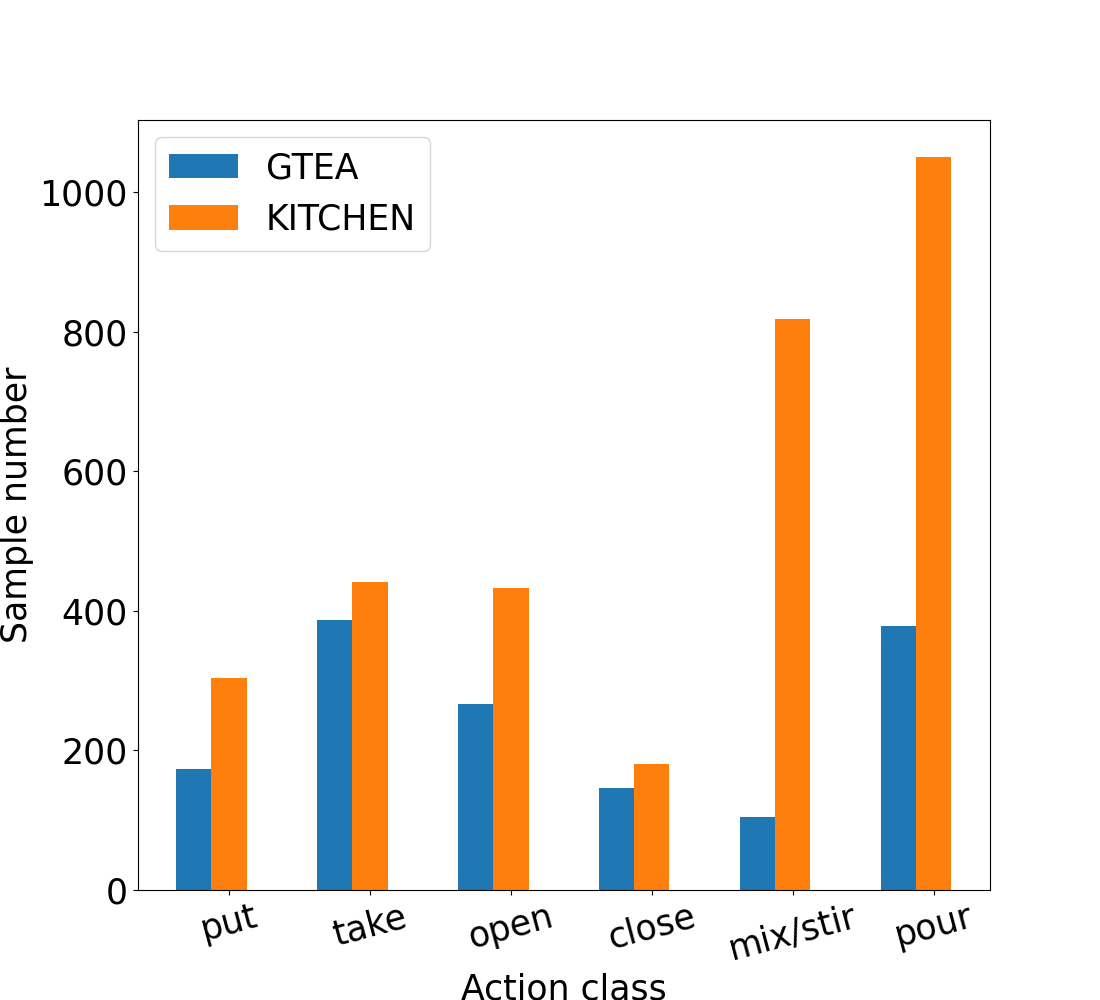}
        \label{fig:gk_class}
    \end{minipage}%
}%
% \centering
%   \setlength{\abovecaptionskip}{-0.2cm}
%   \setlength{\belowcaptionskip}{-0.8cm} 
%   \vspace{-0.2cm}
\vspace{0.8mm}
 \caption{The distribution of classes in the proposed datasets.}
\label{fig:dataset}
\end{figure}

\begin{table*}[t]
%   \label{tab:datasets}
  \centering{
%   \begin{tabular}{p{4.6cm}<{\left}|p{0.8cm}<{\centering}|p{0.8cm}<{\centering}|p{0.8cm}<{\centering}|p{1.1cm}<{\centering}|p{1.1cm}<{\centering}|p{0.9cm}<{\centering}|p{0.9cm}<{\centering}|p{0.9cm}<{\centering}}
\begin{tabular}{p{4.6cm}<{\raggedright}p{0.8cm}<{\centering}p{0.8cm}<{\centering}p{0.8cm}<{\centering}p{1.1cm}<{\centering}p{1.1cm}<{\centering}p{0.9cm}<{\centering}p{0.9cm}<{\centering}p{0.9cm}<{\centering}}
     \toprule
    \textbf{ } & \multicolumn{3}{c}{\textbf{EPIC$_{cvpr20}$}} &  \multicolumn{2}{c}{\textbf{GTEA-KITCHEN}} &  \multicolumn{3}{c}{\textbf{ADL$_{small}$}}\\
   \midrule
    {Resolution} &  \multicolumn{8}{c}{EPIC: 640x480 / GTEA: 456x256 / KITCHEN: 342x256 / ADL: 342x256}  \\
   \midrule
    {Frame rate}  &  \multicolumn{8}{c}{EPIC: 60 / GTEA: 15 / KITCHEN: 30 / ADL: 30}  \\
\midrule
    {Number of classes}  &  \multicolumn{3}{c}{8} & \multicolumn{2}{c}{6}  & \multicolumn{3}{c}{7}\\
\midrule
    {Number of action videos} & \multicolumn{3}{c}{10094} & \multicolumn{2}{c}{454}   & \multicolumn{3}{c}{222} \\
\midrule
    {Domains}  & D1& D2 & D3 &D1 & D2 &D1 &D2&D3\\
% \midrule
    {Number of training segments} &1543 & 2495& 3897&1166 & 2582 &570 & 633 &421\\
% \midrule
    {Number of test segments}  & 435 & 750 & 974 & 291 & 646 & 142 &159 & 106\\
   \bottomrule
%   %\specialrule{0.5pt}{0.5pt}{0.5pt}
  \end{tabular}
 }
  \vspace{4mm}
 \caption{The comparison of the first-person cross-domain video datasets. Note that our proposed datasets adopt different sampling method from EPIC$_{cvpr20}$ due to data augmentation.}
 \label{tab:datasets}
\end{table*}

% There are very few datasets for the first-person video DA dataset except the EPIC-KITCHEN dataset (EPIC), the largest dataset in first-person vision for action recognition and captures daily activities in the kitchen \cite{Damen2020RESCALING}. In this paper, we follow the same setting in \cite{munro20multi} to refer to P08, P01, P22 from EPIC as D1, D2, D3 to build EPIC$_{cvpr20}$ for UDA. Eight verb classes in this dataset for DA are \emph{put}, \emph{take}, \emph{open}, \emph{close}, \emph{mix}, \emph{pour}, \emph{wash} and \emph{cut}. However, EPIC$_{cvpr20}$ is still large scale and may not be very friendly for fast experimentation. Therefore, we propose two small-scale but useful datasets to provide more options to evaluate first-person DA approaches. For both datasets, we follow EPIC$_{cvpr20}$ settings, select the same verb classes from our datasets, and randomly split each class into training and test sets at the ratio of 8:2. ADL$_{small}$ covers the first seven classes, while GTEA-KITCHEN covers the first six, as shown in Figure \ref{fig:dataset}.
The EPIC-KITCHEN dataset (EPIC) is the largest dataset for  first-person action recognition, with daily activities captured in the kitchen \cite{Damen2020RESCALING}. In this paper, we follow the same setting in \cite{munro20multi} to refer to P08, P01, P22 from EPIC as D1, D2, D3 to build EPIC$_{cvpr20}$ for UDA. The eight verb classes in EPIC$_{cvpr20}$ are \emph{put}, \emph{take}, \emph{open}, \emph{close}, \emph{mix}, \emph{pour}, \emph{wash} and \emph{cut}. Despite being a subset of EPIC, EPIC$_{cvpr20}$ is still of large scale, the training in \cite{munro20multi} took 9 hours on 8 NVIDIA V100 GPUs with a batchsize of 128. This sets a high bar for researchers with limited computing resources to do research in this area. Therefore, we propose two small-scale but useful datasets to provide more options to evaluate first-person DA approaches. For both datasets, we follow EPIC$_{cvpr20}$ settings, select the same verb classes from our datasets.
% randomly split each class into training and test sets at the ratio of 8:2.
ADL$_{small}$ covers the first seven classes, while GTEA-KITCHEN covers the first six, as shown in Figure \ref{fig:dataset}.
\begin{figure}[t]
 \centering
\includegraphics[width=0.7 \linewidth]{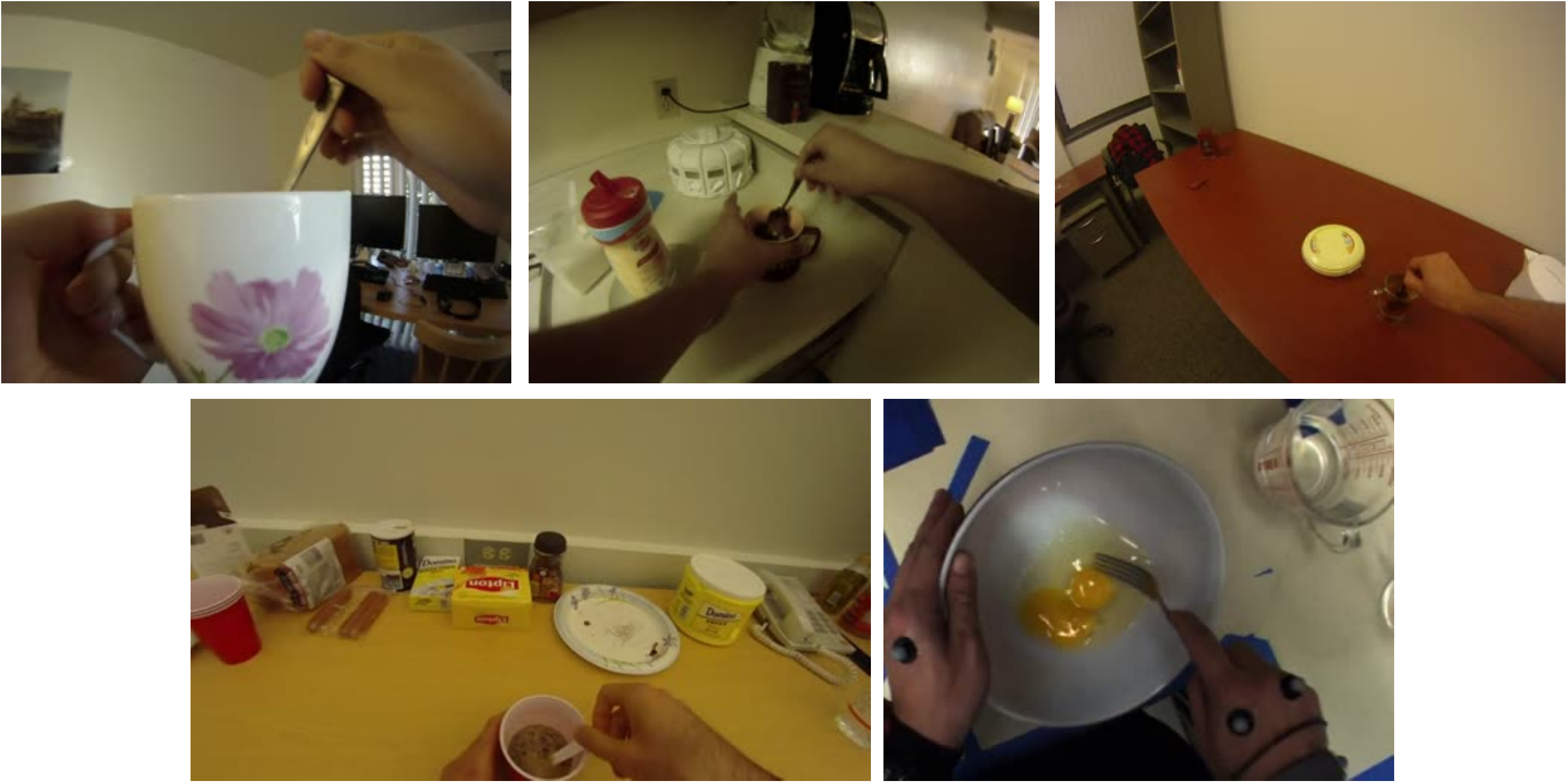}
\vspace{0.8mm}
\caption{Example frames in the class \emph{mix/stir} from the proposed datasets: top: D1, D2 and D3 from ADL$_{small}$; bottom: GTEA and KITCHEN.}
\label{fig:exp_dataset}
\end{figure}

\subsection{ADL$_{small}$} 
\label{sec_adl}
ADL dataset is an activity dataset of daily living in first-person camera views \cite{pirsiavash2012detecting}, containing 10-hour action videos by 20 persons in 20 different apartments. Each video records similar actions by different persons in different environments, which causes significant domain shift
% distinct distributions between videos
as shown in Table \ref{tab:ADL_res}. The original annotation provides action labels. To be consistent with EPIC$_{cvpr20}$, we extract the verbs from action annotations and reorganize them as verb annotations. However, ADL only contains the first seven classes. Therefore, we select videos containing these seven classes. These videos are P4, P6 and P11, which we refer to as D1, D2 and D3 respectively. Considering all the dataset have only 222 action videos, we extract every 16 frames with 4-frame overlap from each action video to make an action sample for data augmentation. After the augmentation, we finally present a new small-scale dataset for UDA on first-person videos, ADL$_{small}$. As the class distribution shown in Fig. \ref{fig:adl_class}, all domains have distinct distribution. Although D3 has all the seven classes, it contains a small number of samples in each class. It is still the smallest after data augmentation. %We keep D3 in our dataset because it can increase the UDA difficulty of our datasets and provide conditions for UDA between small dataset and large dataset. 
% More details are shown in Table \ref{tab:datasets}. 

% In order to provide as many classes for training and evaluation as possible, we select three long-during videos which contain all the seven classes from ADL dataset as our three domains and reorganize the annotations from action to verb categories. These are P4, P6, P11, which we refer to as D1, D2 and D3, respectively. ADL$_{small}$ dataset is the smallest first-person video DA dataset but still has a large domain discrepancy for DA approaches.

\subsection{GTEA-KITCHEN}
GTEA \cite{li2015delving, fathi2011learning} and KITCHEN \cite{de2009guide} datasets are both first-person video datasets recording actions in the kitchen. However, GTEA videos are in the real kitchen, while KITCHEN videos are in a temporary-built kitchen in the lab. Like the dataset name, most actions in these two datasets are about cooking, resulting in six overlapping verbs. We use the whole GTEA and KITCHEN dataset as two domains. As in ADL$_{small}$, we extract the verb annotation from action annotation in GTEA. However, KITCHEN annotation contains many useless segments. Therefore, we re-annotate verbs in KITCHEN to make these two datasets match each other. This leads to another new first-person video dataset for UDA, GTEA-KITCHEN. The class distribution is shown in Figure \ref{fig:gk_class}.% As shown in Table \ref{tab:datasets}, GTEA-KITCHEN has more samples than ADL$_{small}$ but less than EPIC$_{cvpr20}$. 

Table \ref{tab:datasets} summarizes the key statistics for the three datasets. GTEA-KITCHEN has more samples than ADL$_{small}$ but less than EPIC$_{cvpr20}$. For our proposed datasets, we show some sample frames for each dataset in Figure \ref{fig:exp_dataset} with the class \emph{stir/mix}. 
We will study the source-only and target-only recognition accuracy 
% in Table \ref{tab:epicacc} and Table \ref{tab:adlacc}
as well. The target-only setting means training and test are both on the target dataset, while the source-only setting means the model trained on the source dataset is directly tested on the target dataset without UDA. These two results serve as the upper and lower bound of UDA on these datasets.

\section{Experiments}
We evaluate our proposed method on the three datasets: ADL$_{small}$, GTEA-KITCHEN and EPIC$_{cvpr20}$ against other image-based DA networks including DAN \cite{long2015learning}, DANN \cite{ganin2016domain} and CDAN \cite{long2018conditional}. 

\subsection{Experimental Setup}
\textbf{Datasets.}
For EPIC$_{cvpr20}$, we follow the dataset settings in MM-SADA \cite{munro20multi} to split the EPIC$_{cvpr20}$ into training and test sets and randomly sample a 16-frame segment from each action video as the input for both training and test. For our proposed datasets, we follow  similar experimental protocol as EPIC$_{cvpr20}$. We extract every 16-frame segment from each action video as a sample and make adjacent samples to have 4-frame overlap for data augmentation. All the segments are divided randomly into training and test set at a ratio of $8$:$2$, with details in Table \ref{tab:datasets}. 

\textbf{Implementation details.}
We utilize the I3D as our backbone for feature extraction and train all our network end-to-end. Each domain discriminator and classifier are composed of 2 fully connected layers with a dimension of 100 and a ReLU activation function. The only difference from I3D setting is that a dropout rate of 0.5 and a soft-max activation are applied to the classifier to avoid over-fitting and to predict class labels. We select the outputs of the final average pooling layer of I3D, with a dimension of 1024, as the inputs for the discriminator and classifier.

In the training process, we use the labeled source data and unlabeled target data, while in the test process we only utilize the unlabeled target data. For both source and target, the input data is the 16-frame segments sampled from the action video and each frame is resized to 256$\times$256 and randomly cropped to 224$\times$224.
Optimisation is performed using SGD with momentum of 0.9 and batch size of 16. A weight decay with 5e-4 is applied for all parameters. The training process is divided into two stages. First, we set $\lambda_v$ as 0 and train the feature extractor and classifier at a learning rate of 1e-2 for 10 epochs. Second, we follow the same strategy in \cite{ganin2016domain} to increase $\lambda_v$ from 0 to 1 and reduce the learning rate to train the overall network for further 20 epochs. 

% For our proposed datasets, we show the "source-only" and "target-only" recognition accuracy in Table 3 to 5 as well. The "target-only" setting means training and test are both on the target dataset, while the "source-only" setting means the model trained on the source dataset is directly tested on the target dataset without UDA. These two results are regarded to show the upper and lower bound of the dataset for UDA. 

\subsection{Experimental Results}
\textbf{Baselines} Considering that MM-SADA \cite{munro20multi} has a two-stream architecture and $\rm TA^3N$ needs extracted features, we do not include them as our baselines. We extend three state-of-the-art image-based UDA models DAN \cite{long2015learning}, DANN \cite{ganin2016domain} and CDAN \cite{long2018conditional} to videos as our baselines. We follow their default settings except the feature extraction for fair comparison. We conduct each experiment three times with different random seeds and report the average accuracy on test target set as the result for fair comparison. The best result for each task is highlighted in \textbf{bold}, and the second best is \underline{underlined}.

\begin{table}[t]
  \centering{
  \begin{tabular}
  %{c|c|c|c|c|c|c}\\
   {p{1.0cm}<{\centering}p{0.7cm}<{\centering}p{0.7cm}<{\centering}p{0.7cm}<{\centering}p{0.7cm}<{\centering}p{0.7cm}<{\centering}p{0.7cm}<{\centering}}\\
   \toprule
    & Source & DANN & CDAN&DAN& CTAN & Target\\
    \midrule
  {D1$\rightarrow$D2} & 39.4 & 39.4 & \underline{40.7} & 36.3 & \textbf{41.3}  & 52.8 \\
    % \midrule
  {D1$\rightarrow$D3}& 32.0  & 32.9 & 30.3 & \underline{34.1} & \textbf{35.0} & 52.8\\
%   \midrule
%   {D2$\rightarrow$D1}&   &  & &  &  & 60.2\\
%   \midrule
%   {D2$\rightarrow$D3} & 41.6  &  & &  &  & 60.2\\
%   \midrule
%   {D3$\rightarrow$D1}& 39.1  &  & &  &  &  64.7\\
%   \midrule
%   {D3$\rightarrow$D2}&  40.5 &  & &  &  & 64.7\\
%   \midrule
    {Mean} & 35.7  & \underline{36.2} & 35.5 & 35.2 & \textbf{38.1} & 52.8\\
    % \midrule
    {Gain} & - & \underline{0.5} & -0.2 & -0.5 & \textbf{2.4} & -\\
   \bottomrule
  %\specialrule{0.5pt}{0.5pt}{0.5pt}
  \end{tabular}
 }
  \vspace{3.5mm}
 \caption{The comparison of accuracy (\%) with other approaches on EPIC$_{cvpr20}$. On average, we outperform the source-only performance by 2.4\%. Source refers to source-only and Target refers to target-only. Gain represents the absolute difference from the source-only accuracy. The best result for each task is in \textbf{bold}, and the second best is \underline{underlined}.}
 \label{tab:epicacc}
\end{table}

\textbf{EPIC$_{cvpr20}$}
We first evaluate CTAN on 
% by randomly selecting two pair of domains in 
EPIC$_{cvpr20}$ first. We select two hardest UDA tasks D1$\rightarrow$D2 and D1$\rightarrow$D3 according to \cite{munro20multi}. The results are shown in Table \ref{tab:epicacc}. By comparing the recognition accuracy, CTAN outperforms all other baselines on the selected domains. Both CTAN and DANN have improved over the 
% can benefit the UDA performance by comparing with 
the source-only baseline. Specifically, DANN slightly improve the source-only baseline by 0.5\% while CTAN outperforms the baselines by 2.4\% on average, which proves the efficacy of our proposed method. 

% CTAN outperforms all other baselines on the selected domains. Specifically, compared with source-only baselines, CTAN outperforms them by 2.8\% in average. DANN slightly improve the source-only baseline but CTAN outperforms DANN by 2.3\%. It shows that applying our CTA block to DANN can benefit the UDA performance. 

% As mentioned in \ref{sec:adv_uda}, we follow the same stra setting of DANN \cite{ganin2016domain} and outperform DANN by 2.1\% in average.

\textbf{GTEA-KITCHEN}
We then evaluate our network on our proposed GTEA-KITCHEN dataset and the results are shown in Table \ref{tab:gtacc}. For the G$\rightarrow$K task, CTAN outperforms the discrepancy-based approach DAN \cite{long2015learning} by 3.3\%, adversarial-based approaches CDAN \cite{long2018conditional} by 7.1\% and DANN by 2.9\%. For K$\rightarrow$G task, though the performance of CTAN is slightly lower than CDAN, it is still better than other networks. On average, CTAN can achieve the highest accuracy. In comparison, although CDAN performs best in the K$\rightarrow$G task, it gets the worst results (34.0\%) in the G$\rightarrow$K task. 

% CTAN outperforms other approaches by 2.9\% in average, though is not the best in "K-G" task. 

\textbf{ADL$_{small}$}
We finally compare CTAN to the baselines in Table \ref{tab:adlacc}. In general, our network significantly improves on the source-only baseline in 5 out of 6 cases by 2.5\% in average. For D1$\rightarrow$D2 and D1$\rightarrow$D3 tasks, only CTAN achieves better recognition accuracy than the source-only baseline. For D2$\rightarrow$D1 and D2$\rightarrow$D3 tasks, though CTAN did not outperform CDAN (D2$\rightarrow$D1) and DAN (D2$\rightarrow$D3), it still performs much better than DANN (by 3.9\% and 5.1\% respectively) and source-only (by 6.0\% and 4.1\% respectively). 

Overall, CTAN reaches outstanding performance on the three datasets. CTAN has considerably improved the source-only baseline and DANN, which confirmed its efficacy again 
% of our proposed methods: 
% video based UDA and CTA. 
The improvement is consistent for all pairs of domains in 9 out of 10 cases. In contrast, DAN, CDAN and DANN improve the source-only baseline in 5, 6 and 7 out of 10 cases respectively. The only off-target is the D3$\rightarrow$D1 task in ADL$_{small}$, which is reasonable because D3 is the smallest and imbalanced domain among all domains from this dataset as shown in Fig. \ref{fig:adl_class}. In UDA, transferring knowledge from a small dataset with imbalanced classes to a large dataset is still challenging. Similarly, the D3$\rightarrow$D2 task is also improved slightly by CTAN as shown in Table \ref{tab:adlacc}.
%CTAN also improves source-only baseline slightly on D3$\rightarrow$D2 task, as shown in Table \ref{tab:adlacc}.

% We show that our proposed network (CTAN) outperforms discrepancy-based approach DAN \cite{long2015learning} by 2\%, adversarial-based approaches CDAN \cite{long2018conditional} by 0.7\%, as shown in Table \ref{tab:adlacc}. Moreover, our network significantly improves on the source-only baseline in 5 out of 6 cases by 2.5\% in average. For a single case, D1$\rightarrow$D2, all other networks but CTAN fail to achieve better recognition accuracy than source-only baseline.

\begin{table}[t]
  \centering{
  \begin{tabular}
  %{c|c|c|c|c|c|c}\\
   {p{1.0cm}<{\centering}p{0.7cm}<{\centering}p{0.7cm}<{\centering}p{0.7cm}<{\centering}p{0.7cm}<{\centering}p{0.7cm}<{\centering}p{0.7cm}<{\centering}}\\
   \toprule
    & Source & DANN & CDAN&DAN& CTAN & Target\\
\midrule
  {G$\rightarrow$K} & 36.8& \underline{38.2} & 34.0 & 37.8 & \textbf{41.1} &95.9 \\
% \midrule
  {K$\rightarrow$G}& 45.9 & 46.5 & \textbf{48.4} & 43.4 & \underline{47.6} & 94.5\\
%   \midrule
  {Mean}& 41.4 & \underline{42.4} & 41.2 & 40.6 & \textbf{44.3} & 95.2\\
%   \midrule
   {Gain}  & -  & \underline{1.0} & -0.2 & -0.8 & \textbf{2.9} & -\\
   \bottomrule
  %\specialrule{0.5pt}{0.5pt}{0.5pt}
  \end{tabular}
 }
  \vspace{3.5mm}
 \caption{The comparison of accuracy (\%) with other approaches on GTEA-KITCHEN. Source refers to source-only and Target refers to target-only. 
%  The best result for each task is in \textbf{bold}, and the second best is \underline{underlined}.
}
 \label{tab:gtacc}
\end{table}

\subsection{Ablation Study and Analysis}

% \textbf{Domain Availability for Datasets.}
% we explore the domain availability of our proposed datasets. As shown in Table \ref{tab:adlacc}, the average accuracy gap between source-only and target-only results of ADL$_{small}$ is 64.8\%. From Table \ref{tab:gtacc}, the average gap of GTEA-KITCHEN is also over 50\%. Existing UDA approaches, like DAN, DANN, CDAN can improve the performance but not sharply. After adopting CTAN, the gap between target-only result is narrowed to around 50\% but there is still large domain for improving. The result reveals that our proposed small scale dataset is challenging and deserve to be further explored. 
%The average gaps of these two datasets are over 50\% after our proposed CTAN is adopted, which means there is still large domain in these datasets for UDA training. Therefore, although our proposed datasets are small scale, they have large domain discrepancy for exploration.

\textbf{Saturation level of Proposed Datasets.}
We examine the saturation level of our proposed datasets. As shown in Table \ref{tab:adlacc}, the average accuracy gap between source-only and target-only results for ADL$_{small}$ is 64.8\%. From Table \ref{tab:gtacc}, the average gap for GTEA-KITCHEN is  over 50\%. CTAN can narrow the gap by 2.9\% but there is still a large room for further improvement. This shows that our proposed small-scale datasets are challenging and far from saturation so they can support further research in this area. 

\begin{figure}[t]
\subfigure[Target-only]{
    \begin{minipage}[t]{0.5\linewidth}
        \centering
        \includegraphics[width=1.75in]{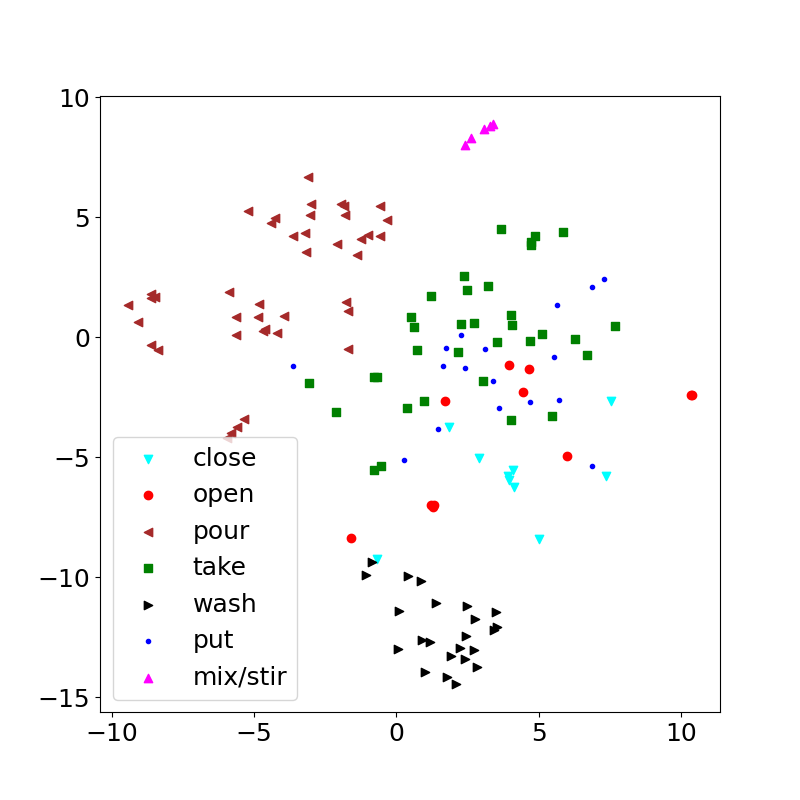}
        \label{fig:tsne_a}
    \end{minipage}%
}%
\subfigure[CTAN]{
    \begin{minipage}[t]{0.5\linewidth}
        \centering
        \includegraphics[width=1.75in]{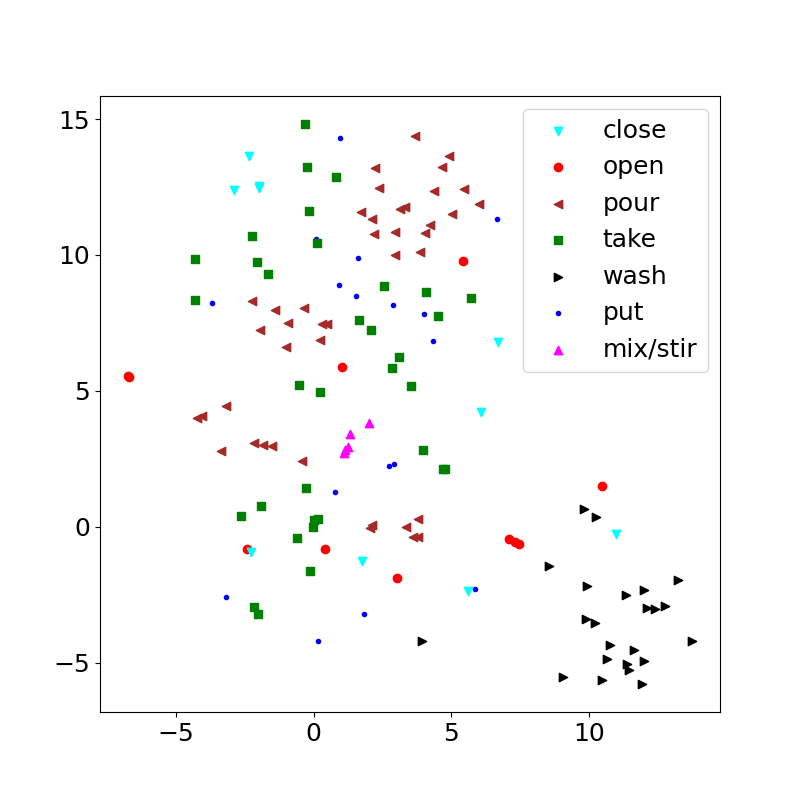}
        \label{fig:tsne_b}
    \end{minipage}%
}%
\centering
  \setlength{\abovecaptionskip}{-0.2cm}
  \setlength{\belowcaptionskip}{-0.8cm} 
  \vspace{0cm}
 \caption{The comparison of $t$-SNE visualization for the feature extractor output on ADL$_{small}$, with D1 as the source domain and D2 as the target domain.}% (Source: D1; target: D2.)}
\label{fig:tsne_compare}
\end{figure}

\textbf{Domain Discrepancy in Datasets.}
We then investigate the reason why the datasets have large discrepancy. We take ADL$_{small}$ as the example and visualize the distribution of feature extractor's last average pooling layer output in both target-only setting and CTAN as shown in Figure \ref{fig:tsne_compare}. As shown, the distributions of the features from classes \emph{pour}, \emph{wash} and \emph{mix/stir} are easy to classify in both Figure \ref{fig:tsne_a} and Figure \ref{fig:tsne_b}. In comparison, the features from the two classes \emph{take} and \emph{put} are mixed together. Unlike \emph{pour}, \emph{wash} and \emph{mix/stir} always occurs with specific objects like kettle or faucet, \emph{take} and \emph{put} can occurs with many objects in many environments. Even in Figure \ref{fig:tsne_a} which is the upper bound performance, the distribution of the features from \emph{take} and \emph{put} classes are not totally separated. It helps us to better understand the challenges of our proposed dataset, and inspires us to tackle it in the future. 
%It reflects the challenge exists in our proposed dataset. Tackling this challenge should be explored in the future.

\begin{table}[t]
  \centering{
  \begin{tabular}
  %{c|c|c|c|c|c|c}\\
   {p{1.0cm}<{\centering}p{0.7cm}<{\centering}p{0.7cm}<{\centering}p{0.7cm}<{\centering}p{0.7cm}<{\centering}p{0.7cm}<{\centering}p{0.7cm}<{\centering}}\\
   \toprule
    & {Source}\centering& {DANN} & {CDAN}& {DAN}& {CTAN} & {Target}\\
\midrule
  {D1$\rightarrow$D2} & 41.1 & 40.6 & \underline{41.1} & 35.9  &\textbf{43.2}  & 95.8\\
% \midrule
  {D1$\rightarrow$D3}& 28.6 & \underline{28.1} & 27.3 & 26.6 & \textbf{31.5} & 95.8\\
%   \midrule
  {D2$\rightarrow$D1}& 25.0 & 27.1 & \textbf{34.0} & 25.7 & \underline{31.0} & 93.5\\
%   \midrule
   {D2$\rightarrow$D3} & 24.8 & 23.8 & 26.2 & \textbf{31.1} & \underline{28.9} & 93.5\\
%   \midrule
   {D3$\rightarrow$D1}& 27.4 & \underline{29.5} & 23.6 & \textbf{31.2} & 26.7 & 95.1\\
%   \midrule
   {D3$\rightarrow$D2}& 37.5 & 37.5 & \textbf{42.7} & 36.5 & \underline{38.2} & 95.1\\
    % \midrule
    {Mean} & 30.7 & 31.1 & \underline{32.5} & 31.2 & \textbf{33.3} &94.8\\
    % \midrule
    {Gain} & - &0.4 & \underline{1.8} & 0.5 & \textbf{2.5}  & - \\
   \bottomrule
  %\specialrule{0.5pt}{0.5pt}{0.5pt}
  \end{tabular}
 }
  \vspace{3.5mm}
 \caption{The comparison of accuracy (\%) with other approaches on ADL$_{small}$. Source refers to source-only and Target refers to target-only.
%  The best result for each task is in \textbf{bold}, and the second best is \underline{underlined}.
}
 \label{tab:adlacc}
\end{table}

\textbf{Inter-dependencies.}
We also measure different variants of CT-block to explore the channel-wise and temporal-wise inter-dependencies. We construct four different blocks. C-block only contains channel-wise attention, while T-block contains only temporal-wise attention. CT-block and TC-block both contain channel-wise and temporal-wise attentions but with different orders. 
%In TC-block, temporal-wise attention is in front of channel-wise attention. 
As shown in Table \ref{tab:adldb}, firstly, CT-block achieves the best accuracy among the four structures and C-block outperforms T-block. It verifies that making network focus on important channels can benefit the feature extraction and UDA. Secondly, channels seem benefit the network more than temporal dimensions because channels carries spatio-temporal information but temporal dimensions do not seem to carry spatial information. Thirdly, the performance of T-block is poorer than the baseline DANN in most pairs. It means simply paying more attention to temporal information may suppress spatial information.

\begin{table}[t]
  \centering{
  \begin{tabular}
  %{c|c|c|c|c|c|c}\\
   {p{0.9cm}<{\centering}p{0.9cm}<{\centering}p{0.9cm}<{\centering}p{0.9cm}<{\centering}p{0.9cm}<{\centering}p{0.9cm}<{\centering}}\\
   \toprule
%   & DANN& C-block&T-block&TC-block&CT-block\\
   & DANN& C$_{block}$&T$_{block}$&TC$_{block}$&CT$_{block}$\\
\midrule
  {D1$\rightarrow$D2} & 40.6 & \textbf{44.8} & 40.9 &40.6 &\underline{43.2}   \\
% \midrule
  {D1$\rightarrow$D3}& 28.1 & 27.6 &27.9 &\underline{29.4} & \textbf{31.5} \\
%   \midrule
  {D2$\rightarrow$D1}& 27.1 & \underline{30.6} & 25.7& 30.3& \textbf{31.0} \\
%   \midrule
   {D2$\rightarrow$D3}& 23.8 & 27.8 &\textbf{28.9} & \textbf{28.9} & \textbf{28.9} \\
%   \midrule
   {D3$\rightarrow$D1}& \underline{29.5} & 19.8 &24.0 & \textbf{31.9} & 26.7 \\
%   \midrule
   {D3$\rightarrow$D2}& 37.5 & \textbf{39.9} &37.2 & 31.3& \underline{38.2} \\
%   \midrule
    {Mean}& 31.1 & 31.8 & 30.8 & \underline{32.1}& \textbf{33.2}\\
    % \midrule
    {Gain} & - - & 0.7 & -0.3 & \underline{1.0} & \textbf{2.1} \\
   \bottomrule
  %\specialrule{0.5pt}{0.5pt}{0.5pt}
  \end{tabular}
 }
 \vspace{3.5mm}
 \caption{The comparison of recognition accuracy (\%) with different blocks on ADL$_{small}$.
%  The best result for each task is in \textbf{bold}, and the second best is \underline{underlined}.
}
 \label{tab:adldb}
\end{table}

% \begin{figure}[t]
% \subfigure[]{
%     \begin{minipage}[t]{0.48\linewidth}
%         \centering
%         \includegraphics[width=1.75in]{tsne_tar_tx.png}
%         \label{fig:tsne_a}
%     \end{minipage}%
% }%
% \subfigure[]{
%     \begin{minipage}[t]{0.48\linewidth}
%         \centering
%         \includegraphics[width=1.75in]{tsne_ct.png}
%         \label{fig:tsne_b}
%     \end{minipage}%
% }%
% \centering
%   \setlength{\abovecaptionskip}{-0.2cm}
%   \setlength{\belowcaptionskip}{-0.8cm} 
%   \vspace{-0.2cm}
%  \caption{The comparison of t-SNE visualization between the output of feature extractor in "target-only" (left panel) and CTAN (right panel) on ADL$_{small}$.(Source is D1 and target is D2.)}
% \label{fig:tsne_compare}
% \end{figure} 

\section{Conclusion and Future Work}
This paper proposed two small-scale action recognition datasets for first-person video domain adaptation, ADL$_{small}$ and GTEA-KITCHEN, both having large domain discrepancy. We utilize these datasets to explore the channel-wise and temporal-wise relationship, and propose channel-wise and temporal-wise excitation attention modules for video to make the network focus on the important channels and temporal dimensions of the CNN features. Finally, we propose Channel-Temporal Attention Network (CTAN) with the attention among channels and temporal dimensions. Our network outperforms the baselines on our proposed small-scale datasets and an existing large-scale dataset. 

Future work can be divided into two categories. First, we will continue to focus on ADL$_{small}$ and GTEA-KITCHEN datasets due to their large domain discrepancy. We will extend more image-based UDA methods for these two datasets and explore the problems about how to recognize \emph{put} and \emph{take}. Second, we plan to extend CTAN to more baselines for robustness test. Considering that CTAN only use video-level alignment, we will also integrate some local alignment methods like channel-wise alignment to make CTAN focus on more common and important information.

% \section{Acknowledgements}
% This work is supported in part by the scholarship from China Scholarship Council (CSC).

% \newpage
{\small
\bibliographystyle{ieee_fullname}
\bibliography{paper.bib}
% \printbibliography
}

\setcounter{figure}{0}
\setcounter{section}{0}

{\Large \textbf{Supplementary Material}}

This supplementary material includes a complete introduction of the datasets and testing time on all three datasets.

\section{Datasets}
This section introduces the details of our small-scale datasets. We follow the similar setting of EPIC$_{cvpr20}$ \cite{munro20multi} to create the datasets. 
% EPIC$_{cvpr20}$ has 8 categories, which are \emph{put}, \emph{take}, \emph{open}, \emph{close}, \emph{mix}, \emph{pour}, \emph{wash} and \emph{cut}. 
Note that \emph{mix} and \emph{stir} are both categorised into \emph{mix} in EPIC$_{cvpr20}$ and we keep the same setting.

\subsection{ADL$_{small}$}
\label{sec:sADL}

We created the ADL$_{small}$ dataset by collecting three videos from the original ADL dataset \cite{pirsiavash2012detecting}, which are P4, P6 and P11. All the videos record real daily life activity. The total length of the videos is one hour and 22 minutes. We restructured the verb annotation in \cite{singh2016first} by removing unclear and non-overlapping labels and segmented all the untrimmed videos into action video clips according to the annotation. We selected seven categories from these videos, which are \emph{put}, \emph{take}, \emph{open}, \emph{close}, \emph{mix}, \emph{pour} and \emph{wash}. The minimum length of each action video is one second, while the maximum is 46 seconds. After restructuring, we extracted every 16 frames with a 4-frame overlap from each action video as the action segment. We then split all the action segments into training and test sets equidistantly in each category with a ratio of 8:2. In the training process, we also split the training set into training and validation sets randomly with a ratio of 9:1. 

As shown in Table \ref{tab:ADL_category}, ADL$_{small}$ contains three domains: D1, D2 and D3, which refer to P4, P6 and P11, respectively. D1 includes 570 training segments and 142 test segments. D2 includes 633 training segments and 159 test segments. D3 includes 421 segments for training and 106 segments for test. ADL$_{small}$ includes 222 original action videos and 2031 extracted action segments in total. ADL$_{small}$ has two characteristics. First, compared with EPIC$_{cvpr20}$, ADL$_{small}$ is about daily life rather than cooking. This leads to 
a larger difference between actions in the same category. For example, \emph{put toothpaste on toothbrush} and \emph{put computer on table} both belong to \emph{put}. Second, some categories are imbalanced. \emph{mix} in D2 and \emph{wash} in D3 have only one action video. This increases the difficulty of UDA on this dataset because the designed network needs to learn the common feature from only one action video.

\begin{table*}[t]
\centering{
%\begin{small}
%   \begin{tabular}{c c c c c c c c c}
    \begin{tabular}{p{0.6cm}<{\centering} p{2.5cm}<{\raggedright} p{1.0cm}<{\raggedleft} p{1.0cm}<{\raggedleft} p{1.0cm}<{\raggedleft} p{1.0cm}<{\raggedleft} p{1.0cm}<{\raggedleft} p{1.0cm}<{\raggedleft} p{1.0cm}<{\raggedleft} p{1.0cm}<{\raggedleft}}
\toprule
& \multirow{2}{*}{} & \multirow{2}{*}{Total} & \multicolumn{7}{c}{Verb category} \\
\cline{4-10}
& \multirow{2}{*}{} & \multirow{2}{*}{} & \emph{put} & \emph{take} & \emph{open} & \emph{close} & \emph{mix} & \emph{pour} & \emph{wash} \\
% \midrule
% D1 & \multicolumn{8}{c}{}
% \midrule
\hline
\multirow{3}{*}{D1} & Action video & 80 & 9 & 19 & 11 & 9 & 4 & 13 & 10 \\
\multirow{3}{*}{} & Action segment & 712 & 46 & 73 & 45 & 41 & 108 & 137 & 262 \\
\multirow{3}{*}{} & Training segment & 570 & 37 & 58 & 36 & 33 & 86 & 110 & 210 \\
\multirow{3}{*}{} & Test segment & 142 & 9 & 15 & 9 & 8 & 22 & 27 & 52 \\
% \midrule
% \midrule
\hline
\multirow{3}{*}{D2} & Action video & 97 & 22 & 32 & 10 & 12 & 1 & 11 & 3 \\
\multirow{3}{*}{} & Action segment & 792 & 89 & 198 & 59 & 49 & 24 & 252 & 121 \\
\multirow{3}{*}{} & Training segment & 633 & 71 & 158 & 47 & 39 & 19 & 202 & 97 \\
\multirow{3}{*}{} & Test segment & 159 & 18 & 40 & 12 & 10 & 5 & 50 & 24 \\
% \midrule
% \midrule
\hline
\multirow{3}{*}{D3} & Action video & 45 & 3 & 20 & 7 & 6 & 2 & 5 & 1 \\
\multirow{3}{*}{} & Action segment & 527 & 42 & 164 & 50 & 58 & 109 & 65 & 39 \\
\multirow{3}{*}{} & Training segment & 421 & 34 & 131 & 40 & 46 & 87 & 52 & 31 \\
\multirow{3}{*}{} & Test segment & 106 & 8 & 33 & 10 & 12 & 22 & 13 & 8 \\
\bottomrule
\end{tabular}
%\end{small}
 \vspace{1mm}
    \caption{The summary of ADL$_{small}$ dataset.}
    \label{tab:ADL_category}
    }
 \vspace{-1mm}
\end{table*}

\subsection{GTEA-KITCHEN}
\label{sec:sGK}

\begin{table}[t]
\vspace{-3mm}
\centering{
%\begin{small}
%   \begin{tabular}{c c}
    % \begin{tabular}{m{1.4cm}<{\centering} | p{5cm}<{\raggedright}}
    \begin{tabular}{m{1.4cm}<{\centering} | m{6cm}<{\raggedright}}
\toprule
GTEA & 
S1\_Cheese\_C1, \hspace{2em} S1\_Coffee\_C1, 
S1\_CofHoney\_C1, \hspace{0.7em} S1\_Hotdog\_C1,
S1\_Pealate\_C1, \hspace{2em} S1\_Peanut\_C1,
S1\_Tea\_C1, \hspace{3.4em} S2\_Cheese\_C1,
S2\_Coffee\_C1, \hspace{2.2em} S2\_CofHoney\_C1,
S2\_Hotdog\_C1, \hspace{1.9em} S2\_Pealate\_C1,
S2\_Peanut\_C1, \hspace{2.2em} S2\_Tea\_C1,
S3\_Cheese\_C1, \hspace{2em} S3\_Coffee\_C1,
S3\_CofHoney\_C1, \hspace{0.7em} S3\_Hotdog\_C1,
S3\_Pealate\_C1, \hspace{2em} S3\_Peanut\_C1,
S3\_Tea\_C1 \\
% \midrule
\hline
KITCHEN & 
S07\_Brownie, \hspace{2.5em} S09\_Brownie,
S12\_Brownie, \hspace{2.5em} S13\_Brownie,
S14\_Brownie, \hspace{2.5em} S16\_Brownie,
S17\_Brownie, \hspace{2.5em} S18\_Brownie,
S19\_Brownie, \hspace{2.5em} S20\_Brownie,
S22\_Brownie, \hspace{2.5em} S24\_Brownie \\
\bottomrule
\end{tabular}
%\end{small}
 \vspace{-1mm}
    \caption{The lists of all collected videos in GTEA and KITCHEN dataset.}
    \label{tab:GK_videos}
    }
\vspace{-3mm}
\end{table}

\begin{figure}[t]
\subfigure[Example frames of GTEA dataset.]{
    \begin{minipage}[t]{0.99\linewidth}
        \centering
        \includegraphics[width=0.99\linewidth]{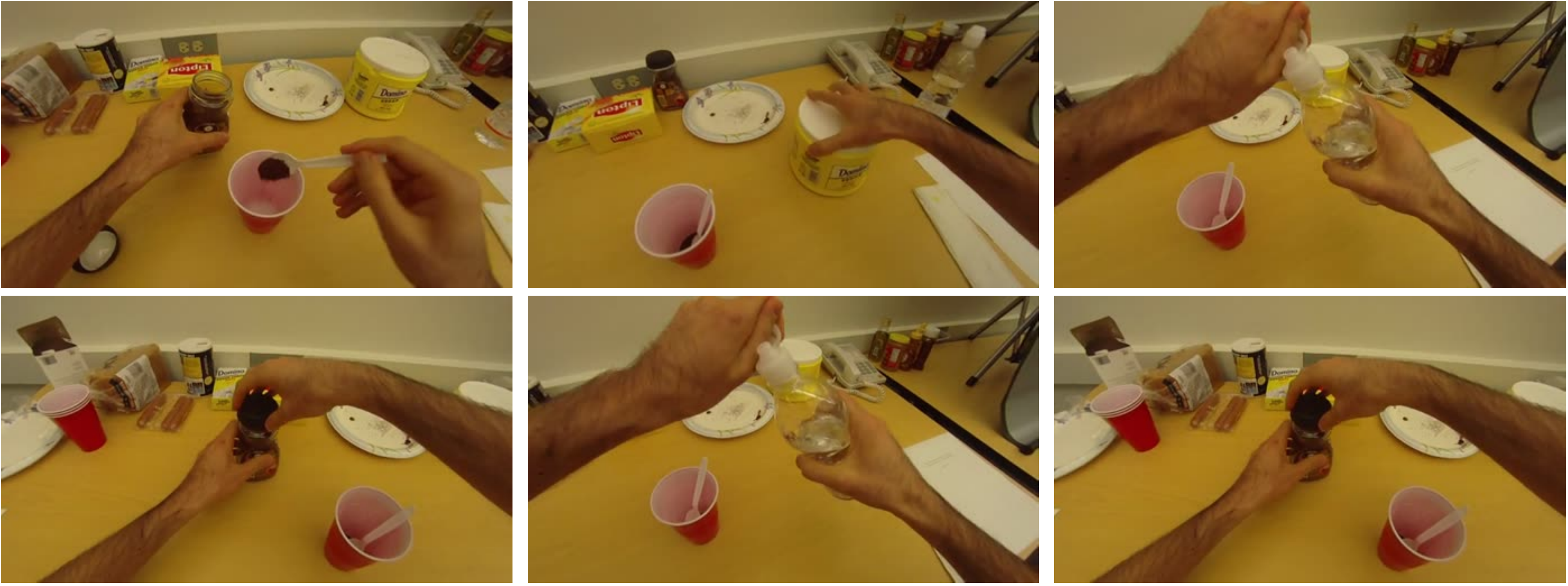} \\
        \label{fig:gtea-a}
    \end{minipage}%
}
\subfigure[Example frames of KITCHEN dataset.]{
    \begin{minipage}[t]{0.99\linewidth}
        \centering
        \includegraphics[width=0.99\linewidth]{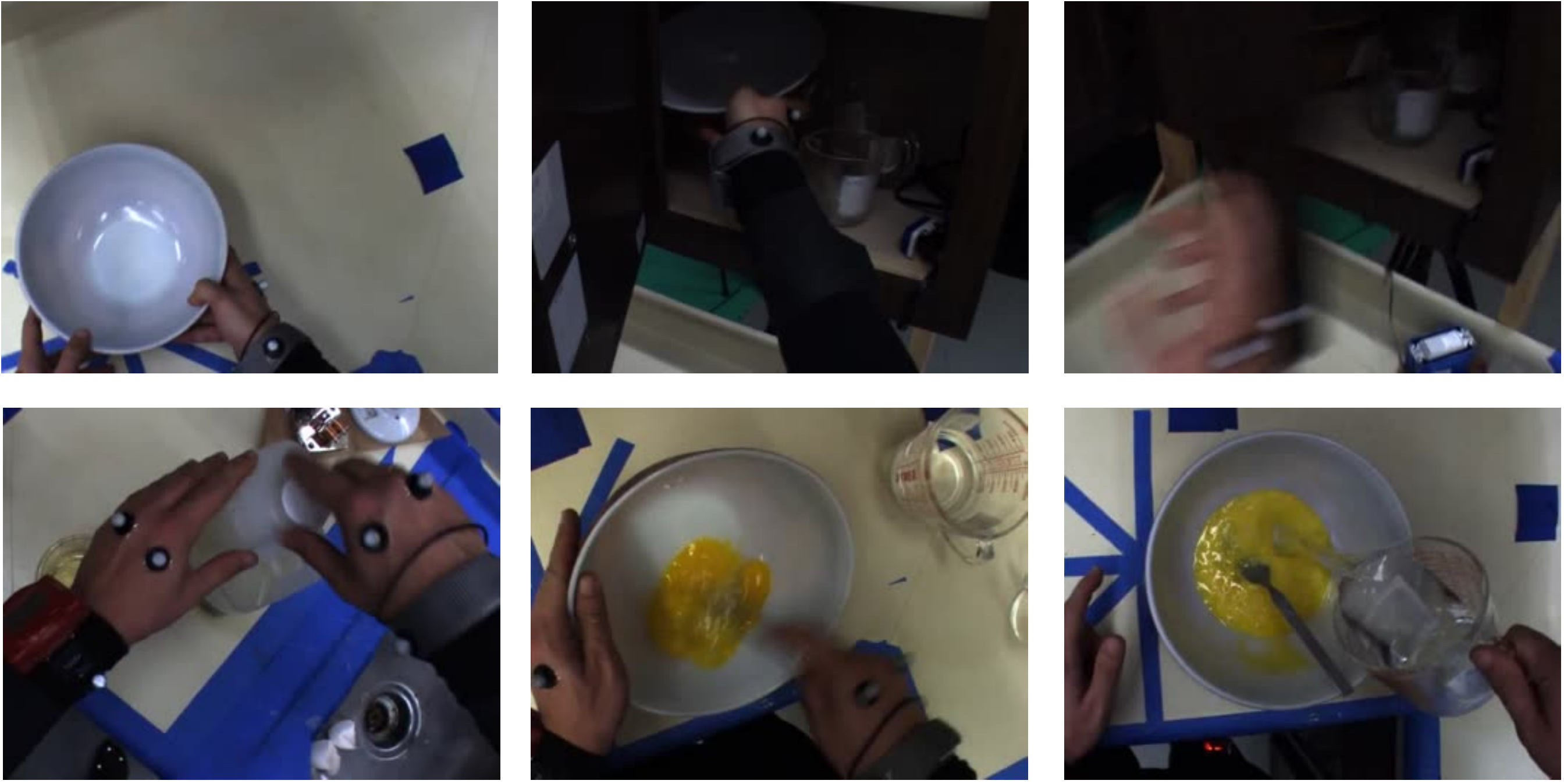}
        \label{fig:kitchen-b}
    \end{minipage}%
}
\centering
    \setlength{\abovecaptionskip}{-8mm}
    \setlength{\belowcaptionskip}{-8mm} 
    \vspace{-2mm}
\caption{Example frames of all categories in GTEA-KITCHEN dataset. From left to right: \emph{put}, \emph{take}, \emph{open}, \emph{close}, \emph{mix}, \emph{pour}.}
\label{fig:gk_example}
\vspace{-3mm}
\end{figure}

\begin{table}[t]
\vspace{-3mm}
\centering{
%\begin{small}
%   \begin{tabular}{c c}
    \begin{tabular}{m{1.5cm}<{\centering} | m{5.9cm}<{\raggedright}}
\toprule
GTEA-KITCHEN & \makecell[c]{GTEA} \\
% \midrule
\hline
\emph{put} &  
\emph{put cheese}, \emph{put mayonnaise}, \emph{put mustard}, \emph{put bread}, \emph{put coffee}, \emph{put sugar}, \emph{put water}, \emph{put honey}, \emph{put hotdog}, \emph{put ketchup}, \emph{put jam}, \emph{put peanut}, \emph{put tea} \\
% \midrule
\hline
\emph{take} & \emph{take bread}, \emph{take cheese}, \emph{take mayonnaise}, \emph{take mustard}, \emph{take cup}, \emph{take coffee}, \emph{take spoon}, \emph{take sugar}, \emph{take water}, \emph{take honey}, \emph{take hotdog}, \emph{take jam}, \emph{take peanut} \\
% \midrule
\hline
\emph{open} & \emph{open cheese}, \emph{open mayonnaise}, \emph{open mustard}, \emph{open coffee}, \emph{open sugar}, \emph{open water}, \emph{open ketchup}, \emph{open honey}, \emph{open chocolate}, \emph{open peanut}, \emph{open tea} \\
% \midrule
\hline
\emph{close} & \emph{close mayonnaise}, \emph{close mustard}, \emph{close coffee}, \emph{close sugar}, \emph{close water}, \emph{close honey}, \emph{close jam}, \emph{close chocolate} \\
% \midrule
\hline
\emph{mix} & \emph{stir spoon}, \emph{stir cup} \\
% \midrule
\hline
\emph{pour} & \emph{pour mayonnaise}, \emph{pour mustard}, \emph{pour water}, \emph{pour coffee}, \emph{pour sugar}, \emph{pour ketchup}, \emph{pour chocolate}, \emph{pour honey} \\
\bottomrule
\end{tabular}
%\end{small}
 \vspace{-1mm}
    \caption{The lists of all collected verb categories in GTEA dataset.}
    \label{tab:G_verb_action}    
    }
\vspace{-3mm}
\end{table}

\begin{table*}[t]
\centering{
%\begin{small}
%   \begin{tabular}{c c c c c c c c c}
    \begin{tabular}{p{1.6cm}<{\centering} p{2.5cm}<{\raggedright} p{1.0cm}<{\raggedleft} p{1.0cm}<{\raggedleft} p{1.0cm}<{\raggedleft} p{1.0cm}<{\raggedleft} p{1.0cm}<{\raggedleft} p{1.0cm}<{\raggedleft} p{1.0cm}<{\raggedleft}}
\toprule
& \multirow{2}{*}{} & \multirow{2}{*}{Total} & \multicolumn{6}{c}{Verb category} \\
\cline{4-9}
& \multirow{2}{*}{} & \multirow{2}{*}{} & \emph{put} & \emph{take} & \emph{open} & \emph{close} & \emph{mix} & \emph{pour} \\
% \midrule
% D1 & \multicolumn{8}{c}{}
% \midrule
\hline
\multirow{3}{*}{GTEA} & Action video & 115 & 20 & 31 & 16 & 15 & 17 & 16 \\
\multirow{3}{*}{} & Action segment & 1457 & 174 & 387 & 267 & 146 & 105 & 378 \\
\multirow{3}{*}{} & Training segment & 1166 & 139 & 310 & 214 & 117 & 84 & 302 \\
\multirow{3}{*}{} & Test segment & 291 & 35 & 77 & 53 & 29 & 21 & 76 \\
% \midrule
% \midrule
\hline
\multirow{3}{*}{KITCHEN} & Action video & 339 & 39 & 109 & 49 & 15 & 51 & 76\\
\multirow{3}{*}{} & Action segment & 3228 & 304 & 442 & 433 & 180 & 818 & 1051 \\
\multirow{3}{*}{} & Training segment & 2582 & 243 & 354 & 346 & 144 & 654 & 841 \\
\multirow{3}{*}{} & Test segment & 646 & 61 & 88 & 87 & 36 & 164 & 210 \\
\bottomrule
\end{tabular}
%\end{small}
  \vspace{1mm}
    \caption{The summary of GTEA-KITCHEN dataset.}
    \label{tab:GK_category}
}
\vspace{-3mm}
\end{table*}

We first collected all 28 videos in the original GTEA dataset \cite{fathi2011learning}, as shown in Table \ref{tab:GK_videos}. The total length of the videos is around 35 minutes. The minimum length of each action video is one second, while the maximum is about 10 seconds. Note that all videos in GTEA are continuous actions without interruption. Therefore, GTEA has shorter video length but more action videos. We extracted verb from action category and collected all of the relevant and overlapping verb categories between GTEA and EPIC$_{cvpr20}$, which results in six categories: \emph{put}, \emph{take}, \emph{open}, \emph{close}, \emph{mix}, \emph{pour}. Each verb category corresponds to multiple action categories in the original GTEA dataset, as shown in Table \ref{tab:G_verb_action}. We then created action video clips, action segments, training set, validation set and test set with the same setting in ADL$_{small}$, as shown in Table \ref{tab:GK_category}.

We collected 12 videos in the KITCHEN dataset \cite{de2009guide}, as shown in Table \ref{tab:GK_videos}. The total length is about one hour and 36 minutes. The minimum and maximum length are one second and 80 seconds, respectively. We manually annotated the verb category for our dataset according to the overlapping categories of GTEA. We followed the same setting to create training, validation and test sets, as shown in Table \ref{tab:GK_category}. Note that the number of \emph{mix} action segments is 3272, which is larger than the sum of other categories. In our experiment, we randomly selected a quarter of them to make \emph{mix} have similar sample sizes with the second most category \emph{pour}, because we do not explore extremely imbalanced UDA in this paper.

We then built the GTEA-KITCHEN dataset, which is the first first-person video dataset across different datasets for UDA on action recognition, including 1166 training segments and 291 test segments from GTEA, 2582 training segments and 646 test segments from KITCHEN. 
There are three characteristics of GTEA-KITCHEN, which provide more options for UDA network exploration. 
First, the resolution difference between GTEA and KITCHEN is a challenge for UDA. The resolution of GTEA data is 456 $\times$ 256, while the KITCHEN is 342 $\times$ 256, as shown in Figure \ref{fig:gk_example}.
% the clarity difference between GTEA and KITCHEN increases the difficulty of UDA. Videos in GTEA are much clearer than the KITCHEN, as shown in Figure \ref{fig:gk_example}. 
Second, illumination change in KITCHEN increases the difficulty of UDA. The brightness in GTEA is nearly unchanged, while the brightness is sometimes very low in KITCHEN, as shown in Figure \ref{fig:gk_example}.
Third, the extremely imbalanced \emph{mix} category mentioned before leads to another challenge worth studying: class-imbalanced domain adaptation.

\subsection{Implementation details}
First, we downloaded all videos of our datasets from their official website. Occupied spaces are 4GB for ADL$_{small}$, 122MB for GTEA and 887MB for KITCHEN. Second, we extracted frames from videos at their respective sampling rates, which are 15 fps for GTEA and 30 fps for both ADL$_{small}$ and KITCHEN. Third, we restructured the annotations following the details in Section \ref{sec:sADL} and \ref{sec:sGK}. 
Finally, we will also publish our annotation files on GitHub.

\section{Testing Time for Our Proposed Datasets}

\begin{figure}[t]
\centering
    \includegraphics[width=\linewidth]{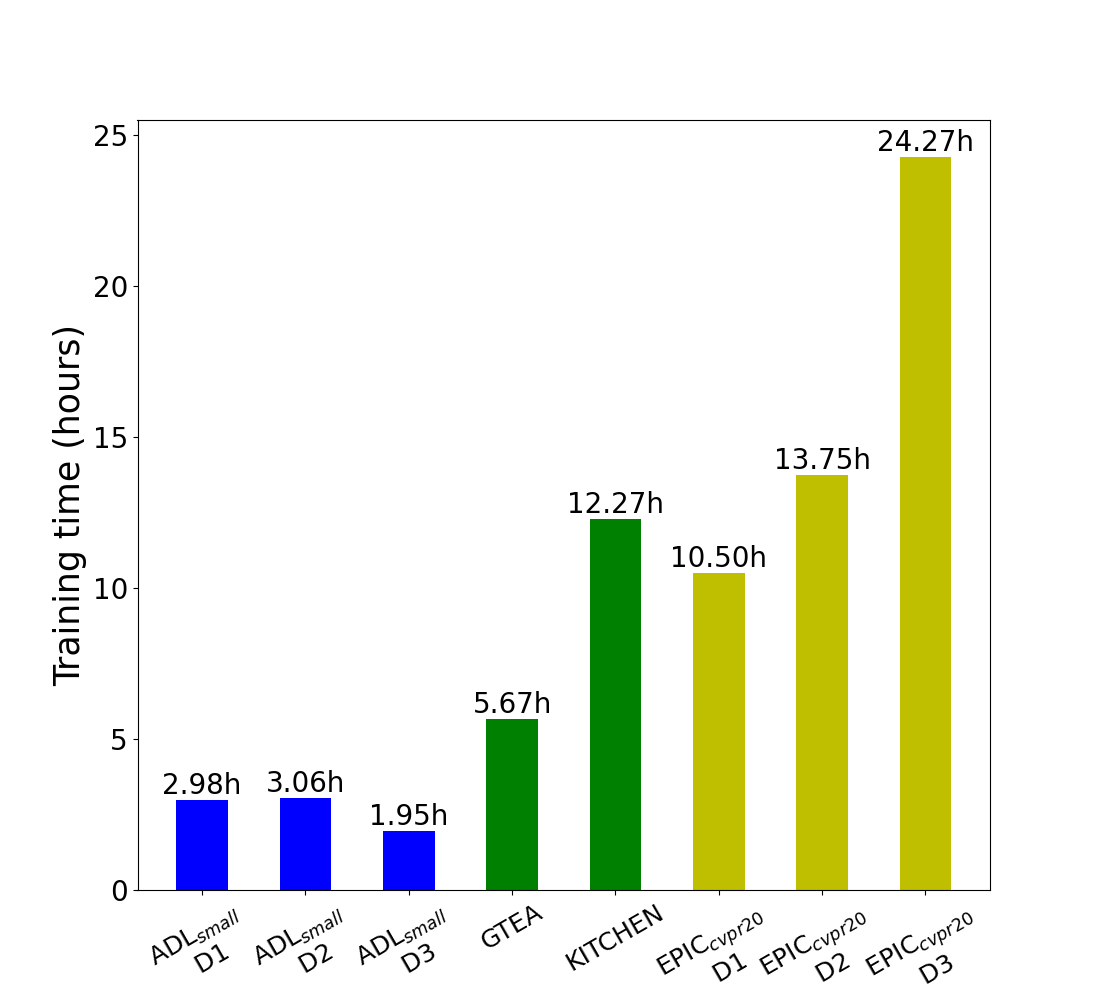}
\vspace{-4mm}
\caption{Training time of source-only setting with 30 epochs on ADL$_{small}$, GTEA-KITCHEN and EPIC$_{cvpr20}$.}
\label{fig:training_time}
\vspace{-3mm}
\end{figure}

Our implementation is based on the PyTorch \cite{paszke2017automatic} and PyTorch Lightning \cite{falcon2019pytorch} framework. We trained the DANN \cite{ganin2016domain} baseline with the source-only setting for 30 epochs on one V100 GPU and reported the time cost. As shown in Figure \ref{fig:training_time}, training takes about three hours on ADL$_{small}$, and nine hours on GTEA-KITCHEN, on average. On EPIC$_{cvpr20}$, the average time is over 16 hours. Note that we utilize full sampling to extract action segments from action videos in our datasets while utilize random sampling to extract only one segment from each action video in EPIC$_{cvpr20}$. Therefore, the time consumed will be reduced more if random sampling is applied to segment extraction in our proposed datasets. This shows our datasets are feasible for researchers with limited computing resource to develop UDA networks.

\vspace{1mm}
% \clearpage
\newpage
{\small
\bibliographystyle{ieee_fullname}
\bibliography{supplement.bib}
}

% \end{document}

\end{document}